\documentclass{egpubl}
\usepackage{egsgp2025}
 
\SpecialIssuePaper         % uncomment for final version of Computer Graphics Forum, special issue
\CGFStandardLicense
\usepackage[T1]{fontenc}
\usepackage{dfadobe}  

\usepackage{cite}  % comment out for biblatex with backend=biber
\BibtexOrBiblatex
\electronicVersion
\PrintedOrElectronic
\ifpdf \usepackage[pdftex]{graphicx} \pdfcompresslevel=9
\else \usepackage[dvips]{graphicx} \fi

\usepackage{egweblnk} 
\usepackage{caption}
\usepackage{booktabs} % For formal tables
\usepackage{hyperref}
\usepackage{xcolor}
\usepackage{subcaption}
\usepackage{wrapfig}
\usepackage{enumitem}
\usepackage{amsmath}
\usepackage{amssymb}

\usepackage{xcolor}
\usepackage{hyperref}
\usepackage{xcolor}
\usepackage{subcaption}
\usepackage{wrapfig}
\usepackage{enumitem}
\usepackage{amsmath}
\usepackage{amssymb}

\usepackage{xcolor}

\title{MDNF: Multi-Diffusion-Nets for Neural Fields on Meshes}

\author[Avigail Cohen Rimon \& Tal Shnitzer \& Mirela Ben Chen
]
{\parbox{\textwidth}{\centering Avigail Cohen Rimon$^{1}$\orcid{0009-0000-7080-6091}, Tal Shnitzer$^{2}$\orcid{0000-0001-9058-1870}, Mirela Ben Chen$^{1}$\orcid{0000-0002-1732-2327} 
        }
        \\
{\parbox{\textwidth}{\centering $^1$Technion - Israel Institute of Technology\\
         $^2$Broad Institute of MIT and Harvard
       }
}
}
\begin{document}

% uncomment for using teaser
% \teaser{
%  \includegraphics[width=0.9\linewidth]{eg_new}
%  \centering
%   \caption{New EG Logo}
% \label{fig:teaser}
%}

\maketitle
%-------------------------------------------------------------------------
\begin{abstract}
We propose a novel framework for representing neural fields on triangle meshes that is multi-resolution across both \textit{spatial} and \textit{frequency} domains.
Inspired by the Neural Fourier Filter Bank (NFFB), our architecture decomposes the spatial and frequency domains by associating finer spatial resolution levels with higher frequency bands, while coarser resolutions are mapped to lower frequencies.
To achieve geometry-aware spatial decomposition we leverage multiple DiffusionNet components, each associated with a different spatial resolution level.
Subsequently, we apply a Fourier feature mapping to encourage finer resolution levels to be associated with higher frequencies.
The final signal is composed in a wavelet-inspired manner using a sine-activated MLP, aggregating higher-frequency signals on top of lower-frequency ones.
Our architecture attains high accuracy in learning complex neural fields and is robust to discontinuities, exponential scale variations of the target field, and mesh modification.
We demonstrate the effectiveness of our approach through its application to diverse neural fields, such as synthetic RGB functions, UV texture coordinates, and vertex normals, illustrating different challenges.
To validate our method, we compare its performance against two alternatives, showcasing the advantages of our multi-resolution architecture.

%-------------------------------------------------------------------------
%  ACM CCS 1998
%  (see https://www.acm.org/publications/computing-classification-system/1998)
% \begin{classification} % according to https://www.acm.org/publications/computing-classification-system/1998
% \CCScat{Computer Graphics}{I.3.3}{Picture/Image Generation}{Line and curve generation}
% \end{classification}
%-------------------------------------------------------------------------
%  ACM CCS 2012
   % (see https://www.acm.org/publications/class-2012)
%The tool at \url{http://dl.acm.org/ccs.cfm} can be used to generate
% CCS codes.
%Example:
\begin{CCSXML}
<ccs2012>
<concept>
<concept_id>10010147.10010371.10010352.10010381</concept_id>
<concept_desc>Computing methodologies~Collision detection</concept_desc>
<concept_significance>300</concept_significance>
</concept>
<concept>
<concept_id>10010583.10010588.10010559</concept_id>
<concept_desc>Hardware~Sensors and actuators</concept_desc>
<concept_significance>300</concept_significance>
</concept>
<concept>
<concept_id>10010583.10010584.10010587</concept_id>
<concept_desc>Hardware~PCB design and layout</concept_desc>
<concept_significance>100</concept_significance>
</concept>
</ccs2012>
\end{CCSXML}

\ccsdesc[300]{Computing methodologies~Machine Learning, Shape Analysis}

\printccsdesc   
\end{abstract}  
%-------------------------------------------------------------------------
\section{Introduction}
Recent advancements in machine learning have lead to a surge of interest in solving visual computing problems using coordinate-based neural networks, known as \textit{Neural fields}. These networks parameterize the physical properties of scenes or objects across spatial and temporal dimensions. Neural fields have gained widespread adoption due to their ability to encode continuous signals over arbitrary dimensions at high resolutions, enabling accurate, high-fidelity, and expressive solutions \cite{xie2022neural}.
They have demonstrated remarkable success in a variety of tasks, including animation of human bodies \cite{he2021arch++}, mesh smoothing and deformations \cite{yang2021geometry, deng2020nasa}, novel view synthesis \cite{mildenhall2021nerf}, mesh geometry and texture editing \cite{yang2022neumesh}, 3D reconstruction \cite{deng2020nasa, shabanov2024banf}, textured 3D reconstruction from images \cite{oechsle2019texture, koestler2022intrinsic}, shape representation and completion \cite{park2019deepsdf}, and neural stylization of meshes \cite{michel2022text2mesh}. %See the recent review about neural fields in visual computing~\cite{xie2022neural}. %\MB{this review is already cited in this paragraph}

Despite their widespread success, these coordinate-based neural architectures remain vulnerable to spectral bias \cite{jacot2018neural} and demand significant computational resources.
Among other generalizations, these shortcomings have been addressed through spatial decomposition strategies using grids \cite{chen2022tensorf, muller2022instant, takikawa2021neural}, which support rapid training and level of detail control.
Additionally, techniques that encode input data using high-dimensional features through frequency transformations, such as sinusoidal representations \cite{mildenhall2021nerf, sitzmann2020implicit, tancik2020fourier}, help mitigate the inherent low-frequency bias of neural fields \cite{tancik2020fourier}.
% Among various generalizations, efforts to make neural fields more efficient and scalable focus on two approaches: spatial decomposition techniques utilizing grids \cite{chen2022tensorf, muller2022instant, takikawa2021neural} that facilitate fast training and level of detail control, and methods that encode input data using high-dimensional feature representations via frequency transformations, such as periodic sinusoidal representations \cite{mildenhall2021nerf, sitzmann2020implicit, tancik2020fourier}, which mitigate the inherent low-frequency bias of neural fields \cite{tancik2020fourier}.

Wu et al. \shortcite{wu2023neural} propose an architecture that bridges these two approaches. They demonstrate that employing different grid resolutions focused on distinct frequency components, combined with proper localization, achieves state-of-the-art performance in terms of model compactness and convergence speed across multiple tasks.
However, the proposed grid-based methods, including \cite{wu2023neural}, are designed for Euclidean spaces and do not account for the unique properties of non-Euclidean, irregular geometric domains like triangle meshes. Although adapting such methods to fit such structures via data modification has shown efficacy, it often overlooks the inherent characteristics of mesh data.
Notably, meshes typically represent smooth manifolds with defined geometry, offering potential for enhanced understanding and representation.
Furthermore, we aim to enhance the architecture's invariance to the multi-representational nature of mesh geometry, accommodating different resolutions and many equivalent vertex connectivities. %, and deformations through rigid motions and isometric transformations for the same geometric structure.
% However, the proposed grid-based methods, including \cite{wu2023neural}, are tailored for Euclidean domains or assume regular data structures.
% Although the adaptation of such methods to irregular geometric domains, such as triangle meshes, is often achieved by modifying the data to fit the existing architecture and has proven effective, this approach disregards the unique properties and structure of mesh data.
% For instance, meshes typically represent smooth manifolds with well-defined geometry, which can be leveraged to enhance the understanding and representation of the mesh structure, enabling accurate, geometry-aware calculations. Furthermore, we want the architecture to be invariant to the multi-representational nature of mesh geometry, which represents the same geometry at various resolutions and features vertices connected in many equivalent ways, deformable by rigid motions and isometries.

Despite significant advancements in learning on meshes, most existing neural architectures primarily prioritize improving generalization capabilities required for classic tasks such as segmentation and classification. Consequently, we find that these approaches struggle when tasked with capturing high-fidelity signals on individual meshes, limiting their effectiveness for neural fields on meshes. Addressing this fundamental gap, our work puts the spotlight on accurately capturing high-resolution signals, setting the foundation for future improvements in intricate signal representation necessary for advanced mesh applications.
% \change{Despite significant advancements in learning on meshes, most existing neural architectures are primarily designed for tasks such as segmentation and classification, where generalization is key and lower-resolution signal representations are sufficient. Consequently, they are less suited for accurately capturing high-resolution signals, limiting their effectiveness for neural fields on meshes.
% This gap underscores the need for architectures that not only leverage the geometric structure of meshes but also effectively model fine-grained signal variations across multiple resolutions.}
% \change{Our work focuses on neural fields, and as such we are mainly interested in overfitting a function to a given mesh. While generalization to a dataset is of outmost importance, if one cannot overfit a high-resolution signal to a single mesh (as we show for existing architectures), then it is quite challenging to design an architecture that can learn such high resolution signals from a dataset. }
% \change{Despite that many works in recent years introduced for learning on triangle meshes, the existing network mostly focus on generalization abilities for tasks as classic tasks such as segmentation and classification requiring lower resolution signals to be represented by the network, while getting harden in representing high resolution signal, which make them less target to this purpose of Neural Fields on meshes.}

In this work, we introduce a novel geometry-aware framework for representing neural fields on triangle meshes that are multi-resolution across both \emph{spatial} and \emph{frequency} domains. Inspired by the Neural Fourier Filter Bank (NFFB) \cite{wu2023neural} and leveraging the geometry-aware DiffusionNet architecture \cite{sharp2022diffusionnet}, our approach decomposes the spatial and frequency domains using multiple DiffusionNet components representing different spatial resolutions and controlling frequency bands using Fourier feature mappings at different scales.
We associate finer spatial resolution levels with higher frequency bands, while coarser resolutions are mapped to lower frequencies. 
This wavelet-inspired decomposition, combined with a carefully designed network architecture, enables our method to effectively learn and represent complex neural fields, accurately capturing intricate details and frequency variations.
We demonstrate the efficacy of our approach through its application to diverse neural fields, including synthetic RGB functions, UV texture coordinates, and vertex normals, showcasing its robustness to discontinuities, exponential scale variations, and mesh modification.
Additionally, we illustrate its practical utility by integrating it into a texture generation model, demonstrating its ability to preserve geometric fidelity and detailed features.

%\vspace{-0.2cm}
% However, effectively capturing the intricate details and frequency variations present in real-world data poses significant challenges for existing neural field architectures.
% While several techniques have been proposed to alleviate the spectral bias inherent in coordinate-based neural networks, enabling them to better represent high-frequency components, many of these approaches are tailored for Euclidean domains or assume regular data structures.
% Adapting these methods to irregular geometric domains, such as triangle meshes, often leads to suboptimal performance or requires intricate modifications.
% Additionally, existing architectures struggle to effectively disentangle and represent the multitude of frequency bands present in complex signals, hampering their ability to accurately capture the rich dynamics observed in real-world data.

\subsection{Related Work}
We highlight relevant work that is related to the key components of our method: architectures for learning on meshes that leverage mesh geometry and neural fields on non-Euclidean domains.

\paragraph*{Learning on Meshes} 
Several works have proposed unique architectures to leverage mesh geometry and other structural properties for learning on meshes \cite{hanocka2019meshcnn,milano2020primal,cohen2019gauge,wiersma2020cnns,yang2020pfcnn,mitchel2021field,smirnov2021hodgenet,sharp2022diffusionnet}. 
Hanocka et al. \shortcite{hanocka2019meshcnn} defined MeshCNN, a convolutional layer on meshes by learning edge features and defining pooling operations through edge collapse. Milano et al. \shortcite{milano2020primal} captures triangle adjacency in meshes through graphs of mesh edges and dual edges.
Cohen et al. \shortcite{cohen2019gauge} and Wiersma et al. \shortcite{wiersma2020cnns} defined equivariant architectures to overcome challenges arising from the mesh geometry. Cohen et al. \shortcite{cohen2019gauge} developed gauge equivariant CNNs on manifolds that depend only on the intrinsic geometry, focusing on signals defined on the surface of the icosahedron, and Wiersma et al. \shortcite{wiersma2020cnns} proposed to overcome the rotational ambiguity of filter kernel transportation by defining rotation-equivariant features for CNNs. Another approach by Yang et al. \shortcite{yang2020pfcnn}, maps surface mesh patches onto flat tangent planes and aligns them to form a flat Euclidean structure, thereby mimicking standard convolutions. 
Mitchel et al. \shortcite{mitchel2021field} define surface convolutions through a scattering operation, which is more resilient to noise due to its aggregation of information from multiple coordinate systems.
Other works use spectral geometry to facilitate learning on meshes. Smirnov et al. \shortcite{smirnov2021hodgenet}, learns spectral geometry elements to construct custom mesh features, and DiffusionNet \cite{sharp2022diffusionnet} leverages the heat equation and learns multiscale diffusion operations to propagate information across the manifold.
These architectures commonly focus on segmentation, classification and correspondence learning tasks. While they form the basic architecture of our work, they are typically incapable of capturing subtle differences in multiple resolutions, as demonstrated in the experimental results, Section \ref{sec:experimental}. %\vspace{-0.2cm}

\paragraph*{Neural Fields}
Neural fields have been increasingly used for learning functional representations in arbitrary resolutions, most commonly for Euclidean domains, e.g. \cite{mildenhall2021nerf,xie2022neural,karras2021alias,park2019deepsdf}. The foundational work, NeRF \cite{mildenhall2021nerf}, a coordinate-based neural network for view synthesis, demonstrated the importance of positional encodings to facilitate learning of high frequency data by neural networks. 
Subsequent works have used periodic activation functions \cite{sitzmann2020implicit}, wavelet-like multi-resolution decomposition \cite{wu2023neural}, and a decomposition to a cascade of band-limited neural fields \cite{shabanov2024banf}.
These works focus on Euclidean spaces, encoding non-Euclidean 2D manifolds as volumes, resulting in higher computational costs or failure to capture the manifold structure. %\vspace{-0.2cm}
\paragraph*{Neural Fields on Manifolds}
Recently, a few works have proposed methods for learning neural fields on non-Euclidean domains \cite{koestler2022intrinsic, xue2023nsf, bensaid2023partial, xiang2021neutex}.
Bensa\"{i}d et al. \shortcite{bensaid2023partial} leverages neural fields for learning partial matching of nonrigid shapes. They use intrinsic positional encodings and a neural representation in the spectral domain to interpolate between matched sparse landmarks of partial shapes. 
NeuTex \cite{xiang2021neutex} represents meshes as 3D volume in a Euclidean space, but encodes texture with a 2D network. To enable texture representation and editing, they train mapping networks between the two spaces, which can be seen as learning a representation of the 2D surface.
Koestler et al. \shortcite{koestler2022intrinsic} takes into account the manifold structure by using the eigenfunctions of the Laplace-Beltrami Operator as positional encodings, serving as point embeddings in the input of the trained neural network. 
This approach is conceptually similar to the concatenation of DiffusionNet \cite{sharp2022diffusionnet} and a Multi-Layer Perceptron (MLP), and we compare our method to such an architecture in Section \ref{sec:experimental}.
Sharing similarities with this approach, Grattarola and Vandergheynst \cite{grattarola2022generalised} propose implicit neural representations based on intrinsic spectral embeddings from the graph Laplacian, generalizing coordinate-based representations to non-Euclidean domains.
% \cite{elhag2023manifold} proposes 
\cite{xue2023nsf} further extends this concept and learns a continuous field, independent of the manifold discretization, by mapping a series of mesh poses to an implicit canonical representation and learning surface deformations fields for each pose. This approach requires a series of related inputs, such as a human mesh in different poses.
Edavamadathil et al. \cite{edavamadathil2024neural} propose Neural Geometry Fields for Meshes, based on partitioning the mesh into parameterizable patches. However, their method is primarily designed for mesh compression, focusing on obtaining a compact neural representation of discrete surface geometry.
A recent work, MeshFeat \cite{mahajan2024meshfeat}, proposes a multi-resolution framework for learning features for neural fields on meshes. Their approach replaces feature learning on Euclidean hash grids with feature learning at mesh vertices across different resolutions, effectively implementing spatial decomposition. However, it aggregates features across resolution levels without explicitly controlling their frequency content, and the learned features remain sensitive to mesh connectivity.

%\vspace{-0.2cm}
\subsection{Contribution}
To summarize, our contributions are as follows:
\begin{itemize}
\item We propose a novel geometry-aware framework for neural fields representation on triangle meshes that is multi-resolution across both \emph{spatial} and \emph{frequency} domains.
% \item Through an illustrative example, we analyze our architecture's pipeline, showcasing the multi-resolution, multi-frequency representations at each stage.
\item We show that our method attains high precision in learning diverse neural fields, such as synthetic RGB functions, UV texture coordinates, and vertex normals, illustrating different challenges.         
\item 
We show that our method outperforms DiffusionNet in learning
high-detail functions, and the NFFB model in handling discontinuous neural fields and mesh modifications.
% We show that our method outperforms DiffusionNet in learning high-detail functions. Moreover, we provide an ablation study comparing our model against a single-resolution variant, demonstrating the efficacy of our multi-resolution approach.
% \item \textcolor{red}{We demonstrate the advantage of our model over the NFFB model in handling discontinuous neural fields.}% and its generalization ability to mesh modifications.}
% \item 
% \change{
% We demonstrate the applicability of our architecture by integrating it into a texture generation model, showcasing its effectiveness in handling detailed geometric features.}
\end{itemize}

\section{Background}
Our architecture draws inspiration from the architecture proposed by Wu et al. \shortcite{wu2023neural}, and builds upon two main works: DiffusionNet \cite{sharp2022diffusionnet} and Fourier feature mapping \cite{tancik2020fourier}. For completeness, we provide here a brief review of these works.
%\vspace{-0.2cm}
\subsection{DiffusionNet}
\label{subsec:back_diffnet}
DiffusionNet~\cite{sharp2022diffusionnet} is a discretization agnostic architecture for deep learning on surfaces.
The architecture consists of successive identical DiffusionNet \emph{blocks}. A central feature of each DiffusionNet block is the use of learned diffusion based on the heat equation to propagate information across the surface.
This diffusion is discretized via the Laplacian $\mathbf{L}$ and mass $\boldsymbol{M}$ matrices of the surface. In our work, we use the cotangent-Laplace matrix, which is ubiquitous in geometry processing applications \cite{crane2013digital, macneal1949solution, pinkall1993computing}.

For efficient diffusion computation, the authors propose to use a spectral method that utilizes the $k$ smallest eigenpairs ${\boldsymbol{\phi}_i, \lambda_i}$ of the generalized eigenvalue problem $\boldsymbol{L} \boldsymbol{\phi}_i = \lambda_i \boldsymbol{M} \boldsymbol{\phi}_i$.
The diffusion layer $h_t(\boldsymbol{u})$ corresponding to time $t$ is implemented by projecting the input feature channel $\boldsymbol{u}$ onto this truncated basis, exponentially scaling the coefficients by $-\lambda_i t$, and projecting back:
\[
    h_t(\boldsymbol{u}) := \boldsymbol{\Phi} 
    \begin{bmatrix}
    e^{-\lambda_1 t} \\
    e^{-\lambda_2 t} \\
    \vdots
    \end{bmatrix}
    \odot (\boldsymbol{\Phi}^T \mathbf{M} \boldsymbol{u})
\]
where $\odot$ denotes the Hadamard (elementwise) product and $\boldsymbol{\Phi}$, $\boldsymbol{\Lambda}$ are the matrices of generalized eigenvectors and eigenvalues, respectively.
The learned diffusion times, optimized per feature channel, control the spatial support ranging from purely local to totally global. Here, we have briefly reviewed only the aspects critical for understanding our method; see \cite{sharp2022diffusionnet} for further details.%\vspace{-0.2cm}
\subsection{Fourier Feature Mapping}
\begin{figure*}[t]
  \centering  \includegraphics[width=0.9\textwidth]{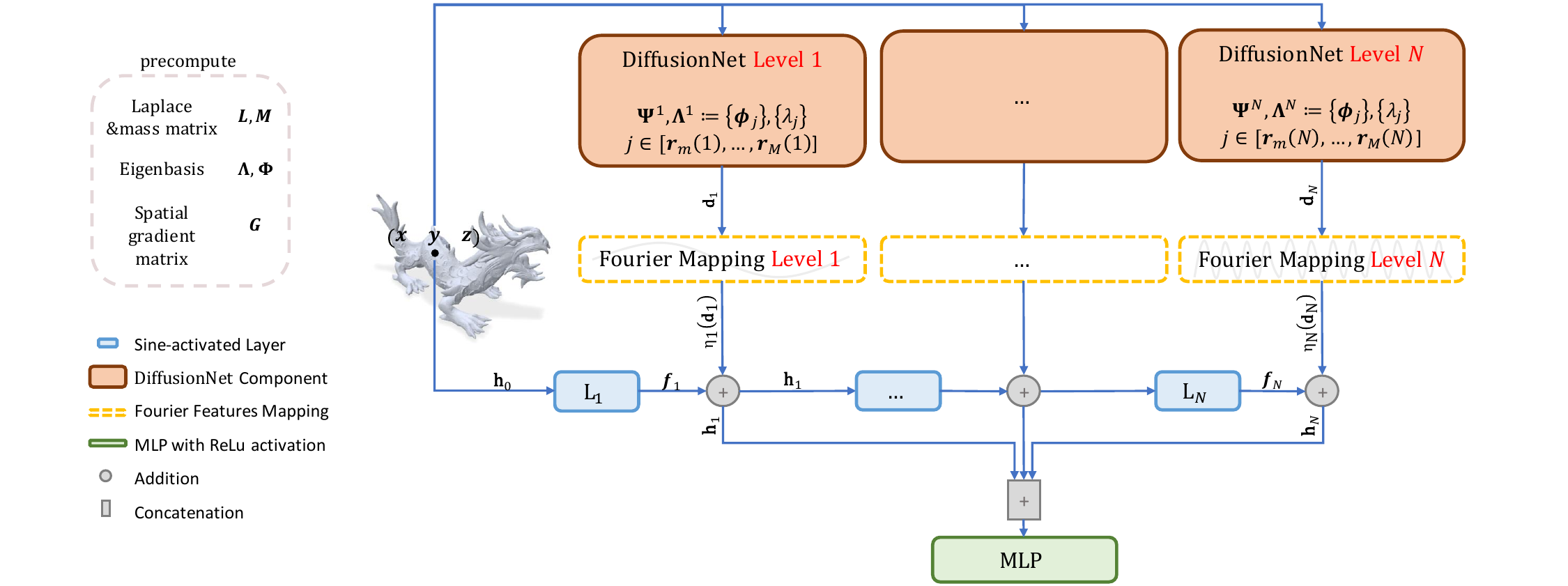}
  \caption{\textbf{Framework overview.} 
    The backbone of our architecture is a sine-activated Multi-Layer Perceptron (MLP), which receives two signals at each linear layer: the output from the previous layer ($\boldsymbol{h}_{i-1}$), representing a low-frequency signal, and another signal produced by the $i$-th resolution level, $\eta_i(\boldsymbol{d}_i)$, representing a higher-frequency signal. The initial input to the first layer $L_1$ is $\boldsymbol{h}_0 := \boldsymbol{X} \in \mathbb{R}^{n\times 3}$. To generate the features $\eta_i(\boldsymbol{d}_i)$, we construct $N$ DiffusionNet components that take $\boldsymbol{X}$ as input and produce $\boldsymbol{d}_i \in \mathbb{R}^{n\times F}$. Fourier Feature mapping layers are then applied to encode these features into appropriate frequencies, resulting in $\eta_i(\boldsymbol{d}_i) \in \mathbb{R}^{n \times m}$. These features are subsequently fed into the linear layer $L_i$ as the higher-frequency components within the sine-activated MLP. To construct the final output, we concatenate the intermediate outputs $\boldsymbol{h}_i \in \mathbb{R}^{n\times m}$ and feed them into a regular MLP with ReLU activation. 
    The definition of $\boldsymbol{G}$
    referenced in the "precompute" rectangle can be found in Supplemental Material Section 1.1.}
  \label{fig:architecture}
\end{figure*}
The work by Tancik et al.~\shortcite{tancik2020fourier} addresses the problem of "spectral bias" in coordinate-based multi-layer perceptrons (MLPs), which refers to their inherent limitation in accurately modeling high-frequency components due to the rapid decay of eigenvalues in their neural tangent kernels (NTKs) \cite{jacot2018neural}.
The authors propose using a Fourier feature mapping that applies a non-linear transformation to the input coordinates before passing them to the MLP.
They report a Gaussian mapping as most effective, where input coordinates are multiplied by random Gaussian matrices to produce high-dimensional Fourier features.
Theoretically, they show that this mapping transforms the NTK into a stationary kernel with a tunable bandwidth. This bandwidth, which determines the width of the kernel’s effective frequency spectrum, is controlled by the scale (standard deviation) of the Gaussian matrices.
A larger scale allows representing higher frequencies, overcoming spectral bias. 

\subsection{Neural Fourier Filter Bank (NFFB)}
Wu et al.~\shortcite{wu2023neural} introduce a novel neural field framework named "Neural Fourier Filter Bank" (NFFB) that decomposes the target signal jointly in the spatial and frequency domains, inspired by wavelet decomposition \cite{shannon1949communication}.
The core idea is to utilize multi-layer perceptrons (MLPs) to implement a low-pass filter by leveraging their inherent frequency bias, and to employ grid features at varying spatial resolutions alongside Fourier feature mappings \cite{tancik2020fourier} at different scales to create a high-pass filter.
A novel aspect of this framework is the correlation of finer spatial resolutions with higher frequency bands, whereas coarser resolutions correspond to lower frequencies. Fourier feature mappings are applied at scales that match the respective spatial resolutions of each grid feature.
The proposed architecture feeds the high-frequency components into sine-activated MLP layers at appropriate depths, mimicking the sequential accumulation of higher frequencies on top of lower frequencies in wavelet filter banks.
This wavelet-inspired decomposition into spatial and frequency components, coupled with the association of specific resolutions with corresponding frequencies, allows for efficient learning of detailed signals while maintaining model compactness and fast convergence.
\section{Method}
\label{sec:method}
We propose a multi-resolution framework that facilitates representing neural fields on meshes across both \textit{spatial} and \textit{frequency} domains. As illustrated in Figure.~\ref{fig:architecture}, our pipeline comprises three key stages: (1) Diffusing features across mesh vertices via multiple DiffusionNet components (Section~\ref{subsec:method_sub1}) to capture spatial variations, (2) Transforming these diffused features through Fourier feature mapping (Section~\ref{subsec:method_sub2}) to associate different frequency bands with the respective resolution levels, and (3) Composing the multi-resolution, multi-frequency signal representation using a sine-activated MLP in a wavelet-inspired manner (Section~\ref{subsec:method_sub3}). We delve into the details of each stage in the following sections.

The input to the network is a matrix $\boldsymbol{X} \in \mathbb{R}^{n \times 3}$, representing the 3D mesh vertex coordinates.
Demonstrating the representational capability of our framework rather than the quality of loss function, the experiments in Section~\ref{sec:experimental} employ supervised learning, where the optimization target is incorporated into the loss function.
We explicitly specify the exact loss formulation used in each experiment.
In Section~\ref{sec:application}, we adopt the configuration from \cite{michel2022text2mesh}, which is unsupervised; see \cite{michel2022text2mesh} for further details.

\begin{wrapfigure}{r}[-5pt]{0.13\textwidth}
\vspace{-15pt}
    \centering    \includegraphics[width=0.13\textwidth]{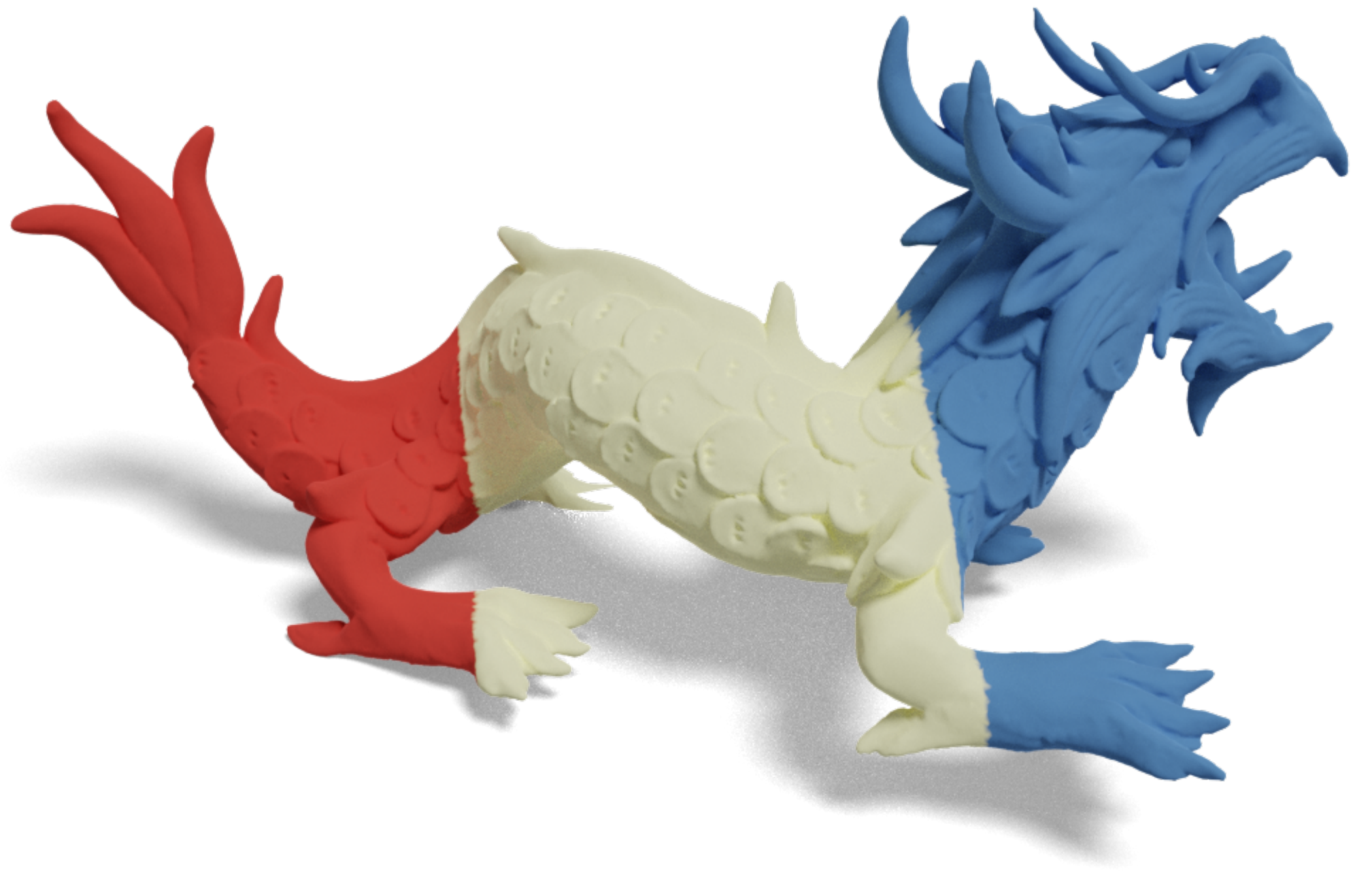}
    \caption*{\textbf{Illustrative example.} Partitioning of mesh vertices.}
\vspace{-15pt}
\end{wrapfigure}

To clarify our architecture's pipeline, we use a synthetic example featuring the Chinese dragon mesh with 125K vertices and 250K faces. The mesh is divided into three groups—Red, Yellowish, and Blue—each linked to a distinct function depicted in the inset figure. These groups represent increasing frequencies: Red corresponds to $\boldsymbol{\phi}_1$, Yellowish to $\boldsymbol{\phi}_{125}$, and Blue to $\boldsymbol{\phi}_{500}$, where $\boldsymbol{\phi}_j \in \mathbb{R}^{n}$ is the $j$-th eigenfunction of the Laplace-Beltrami operator on the mesh, and $n$ is the number of vertices. We generate the target neural field by mapping the patchwork function to an RGB using the HSV colormap.
More details can be found in Subsection~\ref{subsec:exper_1}.
%\vspace{-0.2cm}
\subsection{DiffusionNet Layers}
\label{subsec:method_sub1}
\begin{figure*}[t]
  \centering  \includegraphics[width=0.9\textwidth]{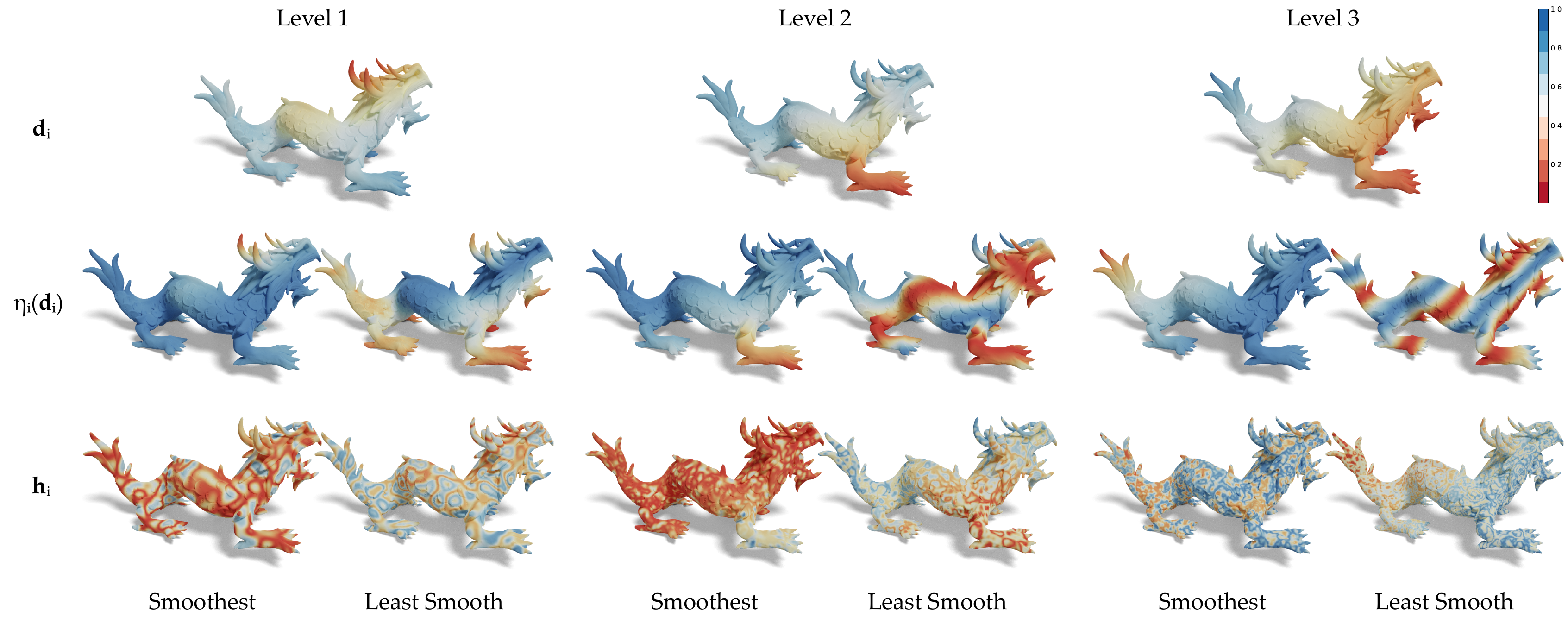}
  \caption{\textbf{Analysis of the illustrative example.} The figure displays the output features at each resolution level across the three key stages of the pipeline. Except for the first stage ($\boldsymbol{d}_i$), where each level contains only one feature ($F=1$), both the smoothest and least smooth features are showcased for each level. Note that the features become less smooth as the stage and level increase.}
  \label{fig:method_dragon1}
\end{figure*}
\begin{figure*}[t]
  \centering  \includegraphics[width=0.9\textwidth]{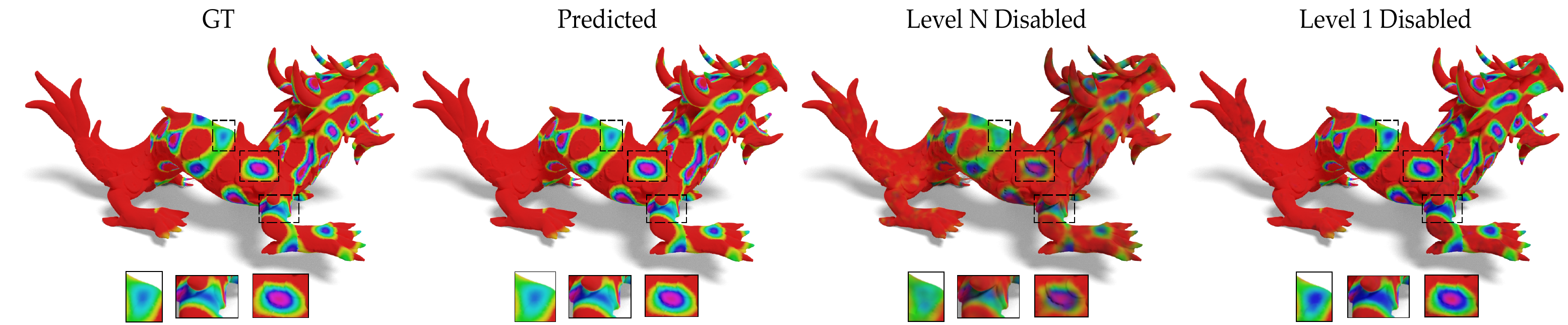}
  \caption{\textbf{Analysis of the illustrative example.} From left to right: (1) Ground truth (GT) RGB function (2) RGB function predicted by our model. (3) Output function with the $N$-th level features $\boldsymbol{h}_N$ disabled. (4) Output function with the first level features $\boldsymbol{h}_1$ disabled. Below each mesh, three zoomed-in areas of the function are presented. Compared to the GT function, the output in (3) appears significantly blurrier, roughly capturing outlines, while the output in (4) presents accentuated contrast, occasionally with an overstated effect.}
  \label{fig:method_dragon2}
\end{figure*}
\paragraph*{Motivation}
% In alignment with the strategy proposed in \cite{wu2023neural}, the first stage of our pipeline involves feeding the input features into a multi-component "layer", where each component aims to store a different resolution band, thereby representing varying spatial resolutions of the mesh features.
% The key distinction between our architecture and NFFB \cite{wu2023neural} lies in substituting the hash grid structure with the DiffusionNet component. As discussed in the background section, the DiffusionNet architecture leverages diffusion layers to enable spatial communication while optimizing the diffusion support per feature channel.
Aligned with the strategy in \cite{wu2023neural}, our pipeline's first stage inputs features into a multi-component "layer," each component associated with a different resolution band, representing varying spatial resolutions of mesh features. Unlike NFFB \cite{wu2023neural}, our architecture replaces each hash grid with a DiffusionNet component. As discussed in Section~\ref{subsec:back_diffnet}, DiffusionNet utilizes diffusion layers to facilitate spatial communication and optimizes diffusion support for each feature channel.

% The choice of adopting the DiffusionNet architecture is based on two main reasons.
% The first stems from the inherent compatibility of hash grid structures with feature representations on regular data structures, such as images or Euclidean spaces. However, our domain of interest revolves around triangle meshes, which exhibit an irregular structure. Although adjustments have been made to adapt triangle meshes to architectures designed for regular-structured data, yielding favorable outcomes \textcolor{red}{[citations needed]}, a geometry-aware approach has demonstrated increased accuracy and efficiency across various applications \textcolor{red}{[citations needed]}.
% As a result, substituting the grid representation with a DiffusionNet architecture intrinsically aligns with the irregular nature of mesh data structures. Furthermore, the DiffusionNet architecture inherently facilitates discretization-agnostic learning \cite{sharp2022diffusionnet}, a desirable property that enhances the generalization capabilities of the overall architecture.
The choice of adopting the DiffusionNet architecture is based on two main reasons.
The first stems from its inherent compatibility with irregular data structures, specifically triangle meshes, as opposed to hash grid structures suited for regular formats like images. Despite previous adaptations of triangle meshes to regular-structured architectures yielding favorable results, a geometry-aware approach like DiffusionNet has proven more accurate and efficient in various applications. Due to that, substituting the grid representation with Diffusionnet is particularly effective for mesh data. Additionally,  DiffusionNet facilitates discretization-agnostic learning, enhancing generalization capabilities of the overall architecture.

% Second, the DiffusionNet architecture intrinsically facilitates a methodology akin to the multi-resolution hash-grid paradigm employed in NFFB.
% The diffusion time parameter, which governs the diffusion support, can be leveraged to enhance a specific spatial resolution through its initialization value assigned to each component.
% Moreover, executing the diffusion process via the suggested "spectral method" enables the association of each DiffusionNet component with a distinct set of eigenvectors.
% This association is also pivotal in effectively controlling the spatial resolution of each component.
Second, DiffusionNet intrinsically facilitates a methodology akin to the multi-resolution hash-grid paradigm of NFFB. The diffusion time parameter can be utilized to adjust spatial resolutions via the initial values assigned to each component. Furthermore, employing the "spectral method" for the diffusion process enables each component to be associated with a distinct set of eigenvectors, enhancing their spatial resolutions as well.

Formally, the DiffusionNet component at the $i$-th level, $\delta_i$, maps the per vertex input 3D coordinate $\boldsymbol{X} \in \mathbb{R}^{n\times 3}$ to an $F$-dimensional feature space:
$\delta_i: \mathbb{R}^{n \times 3} \rightarrow \mathbb{R}^{n\times F}$. 
Let $N$ denote the number of DiffusionNet components. %\vspace{-0.2cm}
\paragraph*{Splitting the spectrum}
Considering the total number of eigenvectors $k_{\text{eig}}$ used for diffusion, we distribute the eigenvectors evenly across the levels, associating the eigenvectors corresponding to the lowest eigenvalues with level 1 and highest to level $N$.

For each level $i \in [1, N]$, we define the range of eigenvector indices used for diffusion in the $i$-th DiffusionNet component as $[\boldsymbol{r}_m(i), \boldsymbol{r}_M(i)]$ where
\begin{equation}
\begin{aligned}
 \boldsymbol{r} := \text{linspace}(0, k_{\text{eig}}, N + 1)   
 \\
 \boldsymbol{r}_m(i) := \boldsymbol{r}(i) \quad \boldsymbol{r}_M(i) := \boldsymbol{r}(i+1) 
\end{aligned}
\end{equation}
where $\text{linspace(\textit{start}, \textit{end}, \textit{steps})}$ is a one-dimensional vector of size \textit{steps} whose values are evenly spaced from \textit{start} to \textit{end}, inclusive.

The corresponding sets of eigenvectors $\boldsymbol{\Psi}_i$ and eigenvalues $\boldsymbol{\Lambda}_i$ used for diffusion at the $i$-th level are:
\begin{equation}
  \boldsymbol{\Psi}_i := \{\boldsymbol{\phi}_j\}_{j=\boldsymbol{r}_m(i)}^{j=\boldsymbol{r}_M(i)}
  \quad \quad
  \boldsymbol{\Lambda}_i := \{\lambda_j\}_{j=\boldsymbol{r}_m(i)}^{j=\boldsymbol{r}_M(i)}
\end{equation}
%\vspace{-0.2cm}
\paragraph*{Splitting diffusion time}
Recall the diffusion time parameter in DiffusionNet controls the spatial resolution of diffusion, theoretically ranging from local to global scales. However, in practice, such range isn't fully realized, as shown in Sec~\ref{sec:experimental}.
The diffusion process, as implemented by the "spectral method", acts as a low-pass filter due to the exponentiation $e^{-\lambda_j t}$, where $t$ is the diffusion time and $\lambda_j$ the $j$-th Laplacian eigenvalue. By creating multiple DiffusionNets, each with distinct eigenvalue ranges, we achieve a refined representation of high-frequency components.

Following the initialization scheme considered in Wu et al~\shortcite{wu2023neural} for the Gaussian distribution
variance (as in our Equation~\eqref{eq:sigma_i_fourier}), we initialize the diffusion times of  the $i$-th DiffusionNet component by $t(i)$ defined as
\begin{equation}
    t(i) := t_{\text{base}} \cdot (t_{\text{exp}})^i
\end{equation} 
Typically $t_{base}$ is set to the squared mean edge length of the mesh and $t_{\text{exp} < 1}$.
% , while 
% is affected by the statistical characteristics of the learned signal.

Figure~\ref{fig:method_dragon1} illustrates the output features of each level at the key pipeline stages, displaying the smoothest and least smooth feature channels per level and stage. See Supplemental Material Section 2 for measuring function smoothness.
The first row shows $\boldsymbol{d}_i := \delta_i(\boldsymbol{x})$ for $i\in[1,2,3]$. Since we set $F=1$ for simplification, only one feature is output at this stage. We observe that the functions at the three levels in this stage exhibit smooth behavior. %\vspace{-0.2cm}
% The second scheme computes $t(i)$ by taking into account the subset of the eigenvectors associated with each component, following the methodology proposed by Sun et al. \cite{sun2009concise}. 
% Considering $k$ eigenvectors for computing the Heat Kernel Signature (HKS) \cite{sun2009concise}, the minimal and maximal diffusion times suggested by the paper are $t_m = \frac{4\ln(10)}{\lambda_k}$ and $t_M = \frac{4\ln(10)}{\lambda_2}$ respectively. Inspired by this, we propose that:
% \begin{equation}
%     t_m(i) := \frac{4\ln(10)}{\lambda_{r_i(1)}}
%     \quad
%     t_M(i) :=
%     \begin{cases}
%     \frac{4\ln(10)}{\lambda_2} & \text{ if } i = 1,\\
%     \frac{4\ln(10)}{\lambda_{r_m(i)}} & \text{ otherwise}.
%     \end{cases}
% \end{equation}

% We then define $t(i)$ by:
% \begin{equation}
% \label{eq:secind_scheme_t}
%     t(i) := t_m(i) + \alpha \cdot (t_M(i) - t_m(i))
% \end{equation}
% where $\alpha \in [0, 1]$ is a hyperparameter.
% The initialization scheme is chosen empirically for each experiment, where the second scheme (Equation~\ref{eq:secind_scheme_t}) facilitates more finely tuned diffusion times.
\subsection{Fourier Feature mapping}
\label{subsec:method_sub2}
\begin{figure*}[t]
  \centering  \includegraphics[width=0.85\textwidth]{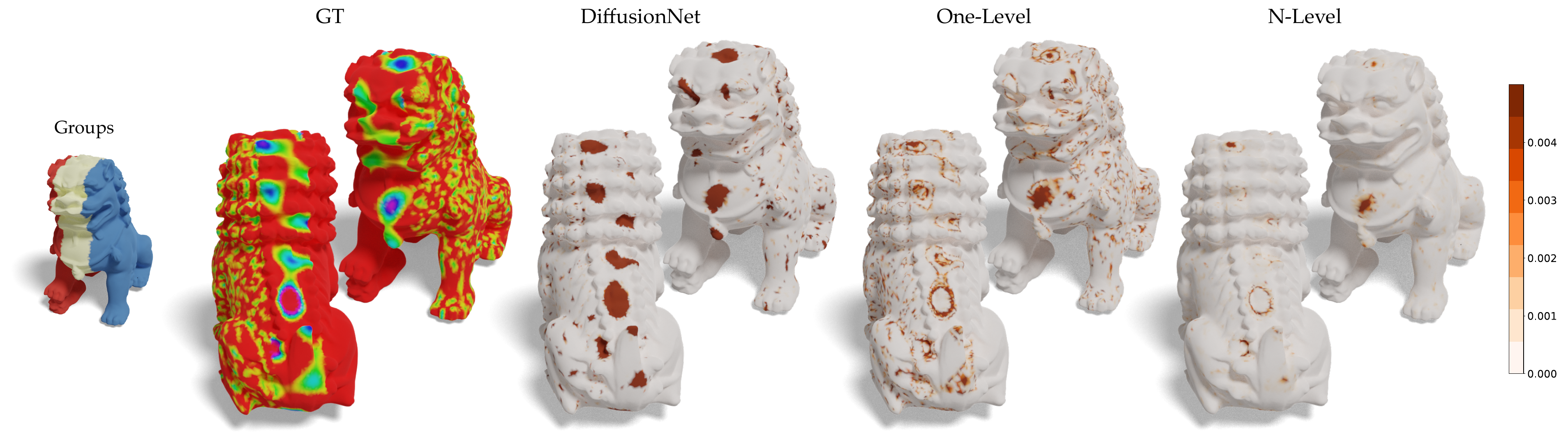}
  \caption{\textbf{Synthetic example results.} From left to right: (1) Partitioning of mesh vertices into three groups, colored Red, Yellowish, and Blue. (2) Two poses of the ground truth RGB function $\boldsymbol{y}_{rgb}$, with each group assigned a distinct function. (3) Error distribution of the field learned by DiffusionNet. (4) Error distribution of the field learned by the One-Level model. (5) Error distribution of the field learned by the N-Level model. Note that our N-Level model demonstrates superior performance relative to the other models, with an error distribution that shows significantly fewer artifacts.}
  \label{fig:synthesis_comparison1}
\end{figure*}
\begin{figure}
  \centering  \includegraphics[width=0.43\textwidth]{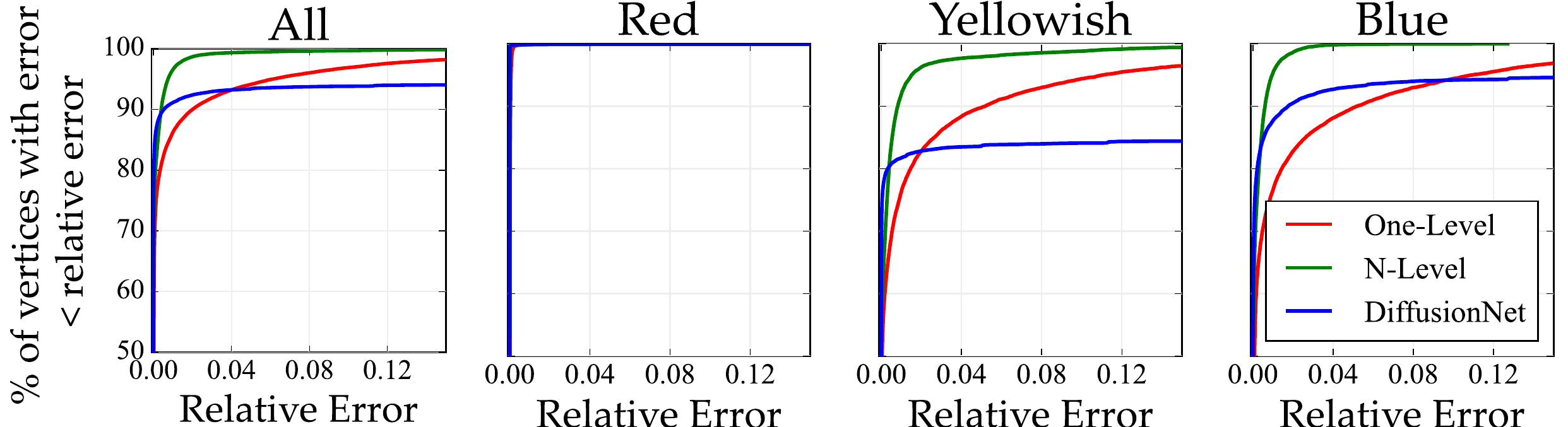}
  \caption{\textbf{Synthetic example results.} Cumulative Distribution Functions (CDFs) of vertex errors for each model across four vertex subsets. We note that for all subsets except the "Red", our N-Level model demonstrates superior performance. For the "Red", the function considered is a constant trivial function which is easily handled by all models. It is noteworthy that although the One-level model requires a longer duration, it ultimately manages to improve, whereas DiffusionNet alone stagnates.}  
  \label{fig:synthesis_comparison2}
\end{figure}
As in NFFB, the Fourier feature mapping stage serves to associate each level of the multi-resolution representation with a distinct frequency band. Inspired by the Fourier feature mapping approach of \cite{tancik2020fourier}, we apply a sinusoidal transformation to the output features from the previous DiffusionNet stage. 

In more details, the Fourier feature at the \(i\)-th level is defined as a mapping from the DiffusionNet output features at the $i$-th level $\boldsymbol{d}_i \in \mathbb{R}^{n \times F}$ to an $m$-dimensional feature space:
{\begin{equation}
    \eta_i(\boldsymbol{d}_i) := \left[ \text{sin}(2\pi  \cdot \boldsymbol{d}_i \cdot \boldsymbol{B}_{i,1}), \dots, \text{sin}(2\pi \cdot 
    \boldsymbol{d}_i \cdot \boldsymbol{B}_{i,m})\right]
\end{equation}
where ${\boldsymbol{B}_{i,1}, \boldsymbol{B}_{i,2}, \dots, \boldsymbol{B}_{i,m}}$ are trainable parameters in $\mathbb{R}^{F}$ forming the frequency transform coefficients on the $i$-th level, and $m$ is a hyper-parameter.
% Crucially, the frequency ranges associated with each level are controlled by the initialization of the $\boldsymbol{B}_{i,j}$ coefficients. Taking inspiration from the Gaussian random Fourier feature mapping proposed by Tancik et al~\shortcite{tancik2020fourier}, we initialize the $\boldsymbol{B}_{i,j}$ using a Gaussian distribution with mean $0$ and level-dependent variance $\sigma_i$ defined in Equation~\ref{eq:sigma_i_fourier}.
% Finer resolution levels (which correspond to higher frequencies) are initialized with larger variance, naturally biasing them towards encoding higher frequency signal components during training. This adaptive initialization scheme allows the multi-resolution representation to automatically disentangle and associate different frequency bands with the appropriate resolution levels without needing to explicitly set fixed frequency ranges a priori.
The frequency ranges for each level are defined by the initialization of the $\boldsymbol{B}_{i,j}$ coefficients. Drawing on the Gaussian random Fourier feature mapping by Tancik et al.~\shortcite{tancik2020fourier}, we set these coefficients using a Gaussian distribution with a mean of $0$ and a level-specific variance $\sigma_i$ (Equation~\eqref{eq:sigma_i_fourier}).
Finer resolutions levels, associated with higher frequencies, are initialized with greater variance, biasing them towards encoding higher frequency signal components. This adaptive initialization approach allows each resolution level to naturally associate with specific frequency bands without pre-setting fixed ranges.

Practically, let $\sigma_{base}, \sigma_{exp} \in \mathbb{R}$, we initialize the $i$-th level coefficients with variance $\sigma_i \in \mathbb{R}$ defined by
\begin{equation}
\label{eq:sigma_i_fourier}
    \sigma_i := \sigma_{base} \cdot (\sigma_{exp})^i
\end{equation}
where $\sigma_{base}, \sigma_{exp}$ are hyper-parameters, and $\sigma_{exp} \geq 1$.

Referring again to Figure~\ref{fig:method_dragon1}, the second row depicts the output features $\eta_i(\boldsymbol{d}_i)$ for $i\in[1,2,3]$.
We observe that the frequency of the least smooth feature increases as the level increases.%\vspace{-0.2cm}

\subsection{Composing the final output}
\label{subsec:method_sub3}
\begin{figure*}[t]
\centering
\begin{subfigure}[b]{\textwidth}
    \includegraphics[width=\textwidth]{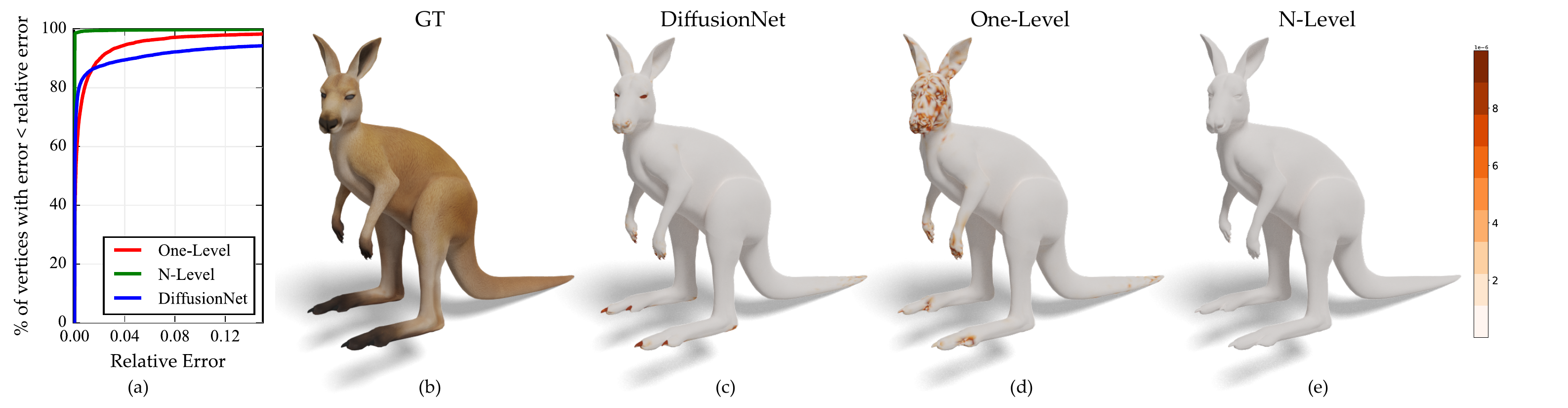}
\end{subfigure}
\begin{subfigure}[b]{\textwidth}
    \includegraphics[width=\textwidth]{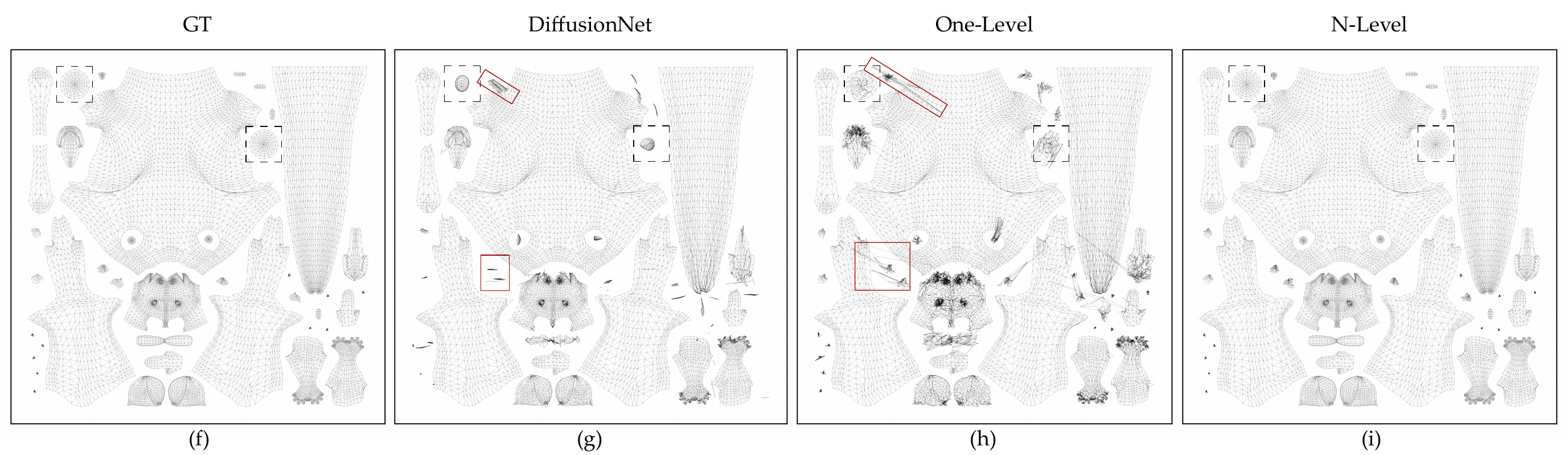}
\end{subfigure}
\caption{\textbf{Discontinuity of mesh and UV coordinates.} Upper row: (a) Cumulative Distribution Functions (CDFs) of vertex errors for each of the three models. (b) Ground truth textured mesh. (c) Error distribution of the UV function learned by the DiffusionNet model. (d) Error distribution of the UV function learned by our One-Level model. (e) Error distribution of the UV function learned by our N-Level model. 
Bottom row: the corresponding learned 2D UV coordinates for each model. Note that our N-Level model (e, i) is free from artifacts at the presented error level and exhibits the best results, also quantitatively, as shown in (a).
Conversly, the DiffusionNet model (c, g) exhibits high errors at the eyes (dashed rectangles in (f, g, h, i)) and other pointed areas, and tends to squash small areas in the UV coordinates.
Our One-Level model shows significant errors in the head region (d), and creates undesirable overlaps in the UV coordinates (h). Examples for high error areas in the UV are marked by red rectangles in (g, h).}
\label{fig:uv_comparison_kangaroo}
\end{figure*}
% The next stage involves composing the final output from the Fourier transformed features $\eta_i(\boldsymbol{d}_i) \in \mathbb{R}^{m}$ across the different levels $i \in [1, N]$.
% Before getting into details, let us start from two key observations from \shortcite{wu2023neural}: First, after neural processing the features across levels are not necessarily orthogonal, calling for learned layers to optimally combine them while mitigating this non-orthogonality.
% Second, implementing residual connections facilitates the aggregation and joint updating of the multi-resolution features by ensuring all levels maintain a similar processing depth within the network.
The next stage composes the final output using the Fourier transformed features $\eta_i(\boldsymbol{d}_i) \in \mathbb{R}^{n \times m}$ across levels $i \in [1, N]$.
Two critical observations from \cite{wu2023neural} inform this process: First, features across levels are not necessarily orthogonal, calling for learned layers for optimal combination and mitigation of non-orthogonality. Second, implementing residual connections helps aggregate and joint update of the multi-resolution features, maintaining consistent processing depth across all levels in the network.

We thus start by applying a sine-activated MLP \cite{sitzmann2020implicit} that takes in the Fourier features $\eta_i(\boldsymbol{d}_i)$ in a manner that sequentially accumulates higher-frequency components on top of lower-frequency components.

More formally, let us denote the $i$-th layer as $L_i$ where $i \in [1, N]$, using $\boldsymbol{f}_i$ to represent the output of $L_i$, and $\boldsymbol{h}_i$ to denote the combination of the $i$-th layer output with the next level's features:
\begin{equation}
\begin{split}
\label{eq:f_i_h_i}
\boldsymbol{f}_i[v]^T := \sin (\alpha_i \cdot \mathbf{W}_i \cdot \boldsymbol{h}_{i-1}[v]^T + \mathbf{b}_i), \quad
\boldsymbol{h}_i := \boldsymbol{f}_i + \eta_i(\mathbf{d}_i)
\end{split}
\end{equation}
where for ease of notation we define $\boldsymbol{h}_0 := \mathbf{X} \in \mathbb{R}^{n\times 3}$.
Since the MLP is weight-shared across vertices, we denote by $\boldsymbol{f}_i[v], \boldsymbol{h}_i[v]$ the entries of $\boldsymbol{f}_i, \boldsymbol{h}_i$ corresponding to vertex v, respectively.
Here, $\mathbf{W}_i \in \mathbb{R}^{m \times m}$ (with $\boldsymbol{W}_0 \in \mathbb{R}^{m\times 3}$) and $\mathbf{b}_i \in \mathbb{R}^{m}$ are the trainable weight and bias parameters in layer $L_i$, and $\alpha_i$ is analogous to the $w_0$ hyperparameter in SIREN \cite{sitzmann2020implicit}, acting as a frequency scaling factor that allows controlling the frequency band that this level focuses on representing.

Next, as illustrated in Figure~\ref{fig:architecture}, we establish residual connections by concatenating the outputs$\boldsymbol{h}_i \in \mathbb{R}^{n\times m}$ from each level and passing them through an additional MLP with ReLU activations, while also transferring them to the subsequent layer $L_{i+1}$ as described earlier.
Alternatively, as suggested in \cite{wu2023neural}, instead of concatenating $\{\boldsymbol{h}_i\}_{i=1}^{N}$, one could pass each feature through a per-level linear layer $O_i$ and sum the outputs to obtain the final feature representation. We refer to \cite{wu2023neural} for further details.

The third row in Figure~\ref{fig:method_dragon1}, representing the output features $\boldsymbol{h}_i$ for $i \in [1,2,3]$, exhibits features that are significantly less smooth than those in previous stages.
Further, in this stage we can see that for both the smoothest and least smooth features, higher levels correspond to increasingly noisier features.

% To gain insight into the resolutions levels learned by the trained network, Figure~\ref{fig:method_dragon2} illustrates the output neural field, which represents the RGB function, during evaluation when disabling the first or the last level.
% We disable the features of level $i$ artificially by setting $\boldsymbol{h}_i = \boldsymbol{0}$.
% Generally speaking, we observe that disabling level $N$ results in a relatively blurry function, corresponding to lower frequency components, while disabling level $1$ yields a high contrast function, associated with higher frequency components.
% To demonstrate these observations, we zoom in on three areas in each of the ground truth (GT) function, the function predicted by our model, the function with level $N$ disabled, and the function with level $1$ disabled.
% Compared to the GT and our predicted function, the output with level $N$ disabled appears significantly blurrier, primarily capturing only the rough outlines of textures.
% Conversely, disabling level $1$ results in high-contrast transitions, sometimes exaggerated.
% For instance, areas that are light blue in the GT and surrounded by similar light green regions appear as a more distinct blue, despite their subtle difference in hue in the GT.
To gain insight into the resolutions levels learned by our trained network, Figure~\ref{fig:method_dragon2} depicts the output neural field representing the RGB function during evaluation, with either the first or last level disabled. Disabling level $N$ generally results in a blurry output, reflecting lower frequency components, whereas disabling level $1$ produces a high-contrast function tied to higher frequencies. We analyze this by zooming in on three areas across the ground truth (GT), our model’s predicted function, and outputs with level $N$ or $1$ disabled. The output from disabling level $N$ is notably blurrier, capturing only basic texture outlines compared to the GT. Conversely, disabling level $1$ enhances contrast, often exaggerating transitions. For instance, areas that are light blue in the GT and surrounded by similar light green regions appear as a more distinct blue, despite their subtle difference in hue in the GT.%\vspace{-0.2cm}

\subsection{Implementation Details}
The implementation details such as the loss function, number of training epochs, and network size are adjusted for each experiment.
Outside of these customized components, the overall training setup remains consistent across all experiments.
We implement our method in PyTorch \cite{paszke2019pytorch}, and utilize the Adam optimizer \cite{kingma2014adam} with the default settings of $\beta_1 = 0.9$ and $\beta_2 = 0.99$.
The learning rate is set to $10^{-4}$, and it is reduced by a factor of 0.7 every 700 iterations. We set the output dimension of DiffusionNet components as $F=2$, and the maximal index of the Laplacian eigenpair considered in the diffusion process, $k_{\text{eig}}$, is set to 500.
We run all experiments on a single NVIDIA A40 GPU.
For brevity, only the essential details are presented here; for a detailed description of the hyperparameters, see the Supplemental Material.

\section{Experimental Results}
\label{sec:experimental}
\begin{figure*}[t]
\centering
\includegraphics[width=\textwidth]{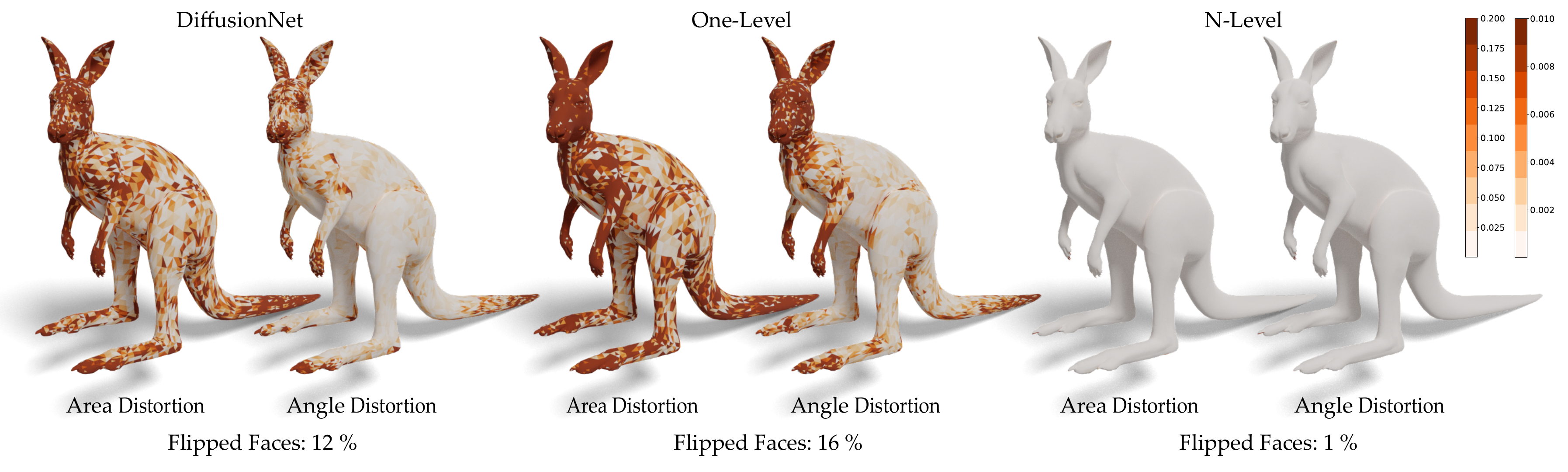}
\caption{\textbf{Discontinuity of mesh and UV coordinates.} Three UV distortion metrics are presented for each model: area distortion, angle distortion, and the percentage of flipped faces. The area and angle distortion are visualized as mesh functions for each model, while the percentages of flipped faces are noted in text in the bottom row. We observe that our N-level model significantly outperforms the two other models in terms of these metrics.}
\label{fig:uv_distortions_kangaroo}
\end{figure*}
\begin{figure}
\centering
\includegraphics[width=\linewidth]{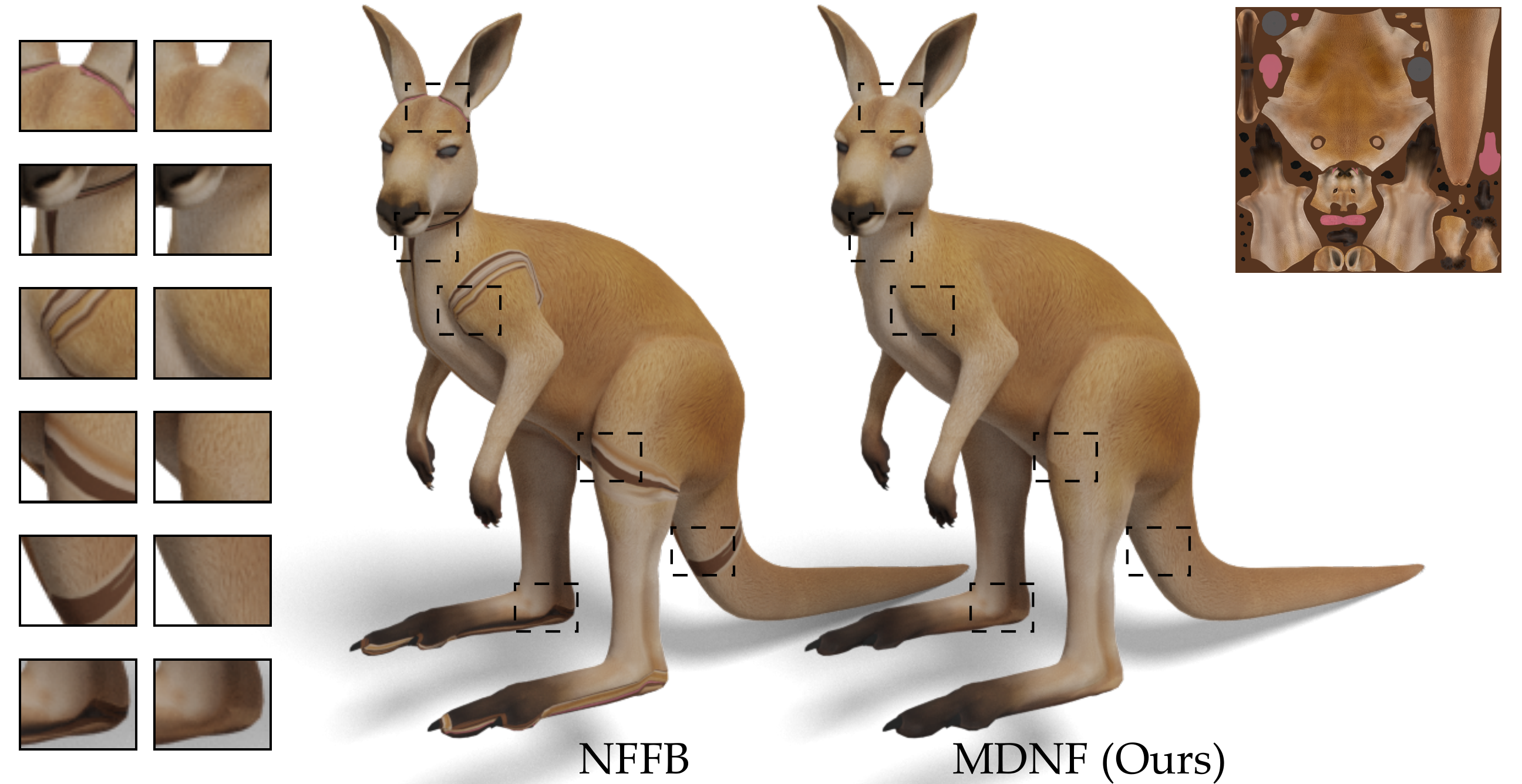}
\caption{\textbf{Discontinuity of mesh and UV coordinates.} Textured mesh visualization using UV coordinates learned by NFFB (left) and our N-Level model (right). Areas of texture mapping discontinuities reveal significant artifacts in the NFFB results, while our method maintains mapping accuracy. Zoomed insets (left) highlight key areas where the differences are most prominent. The texture image is shown in the top right for reference.}
\label{fig:kangaroo-nffb}
\end{figure}
We evaluate our method on three neural fields: synthetic RGB function (Section~\ref{subsec:exper_1}), UV texture coordinates (Section~\ref{subsec:exper_2}), and vertex normals (Section~\ref{subsec:exper_3}).
Focusing on demonstrating our architecture, all models were trained using a supervised approach. 
% As our focus is on demonstrating our architecture rather than the loss functions, all models were trained in a supervised approach.

We compare against two baselines: a single DiffusionNet component, and our method with $N=1$, denoted as the One-Level model.
We refer to our method as the N-Level model where $N>1$, with $N$ determined empirically for each experiment.
For all models, we report the results of the best-performing configuration.%\vspace{-0.2cm}

In Sections~\ref{subsec:exper_2} and~\ref{subsec:exper_3}, we also compare the performance of the NFFB architecture with our N-level model to further motivate our architecture.

\subsection{Synthetic Example}
\label{subsec:exper_1}
To illustrate the effectiveness of our method, we start with a synthetic example, resembling the one in the Section~\ref{sec:method}.
In this experiment we define the target field to be a 3-channel RGB function $\boldsymbol{y}_{rgb} \in \mathbb{R}^{n \times 3}$ defined on a mesh, where $n$ denotes the number of mesh vertices.
We train the model by minimizing the mean square error (MSE), hence our loss function for this task is
\begin{equation}
\mathcal{L}_{\text{rgb}}(\boldsymbol{y}) := \frac{1}{n}
{ \lVert \boldsymbol{y} - \boldsymbol{y}_{rgb} \rVert}_2^2
\end{equation}
% \paragraph{\textbf{Data}}
\textbf{Data.}
We demonstrate this example using the Chinese lion mesh, composed of $50$K vertices.
%\vspace{-0.2cm}
\paragraph*{Neural Field Generation}
% We generate the function $\boldsymbol{y}_{rgb}$ in the following way.
% First, we define a partition of the mesh vertices into 3 groups determined by their $x$ coordinate.
% The partition of the mesh can be seen in Figure~\ref{fig:synthesis_comparison1}, with each group colored differently.
% We denote the different groups by $\text{group}_i$ for $i \in [1,2,3]$, where $\text{group}_1$, $\text{group}_2$, and $\text{group}_3$ correspond to the Red, Yellowish, and Blue groups, respectively.
% We then consider three scalar functions on the mesh vertices; $f_1 = \phi_1 \in \mathbb{R}^{n}$ (which is a constant function), $f_2 = \phi_{125} \in \mathbb{R}^{n}$, and $f_3 = f_p \in \mathbb{R}^{n}$ generated as Perlin noise on the mesh \cite{perlin1985image, vigier_perlin_numpy}.
% Note that the frequency of the functions $f_i$ increases with $i$.
% Let us define a patchwork function $g \in \mathbb{R}^{n}$ on the mesh by $g[\text{group}_i] := f_i[\text{group}_i]$.
% The function $\boldsymbol{y}_{rgb}$ is generated by applying a HSV colormap to the function $g$. First, $g$ is normalized to the range $[0,1]$ and then used to determine the Hue parameter.  See $\boldsymbol{y}_{rgb}$ in the GT figures in  Figure~\ref{fig:synthesis_comparison1}. 
To generate the function $\boldsymbol{y}_{rgb}$, we first partition the mesh vertices into three groups based on their $x$ coordinates, as visualized in Figure~\ref{fig:synthesis_comparison1}. We denote the Red, Yellowish, and Blue, groups by $\text{group}_1$, $\text{group}_2$, and $\text{group}_3$, respectively.
We assign to each group a scalar function: $\boldsymbol{g}_1 := \boldsymbol{\phi}_1 \in \mathbb{R}^{n}$ (constant), $\boldsymbol{g}_2 := \boldsymbol{\phi}_{125} \in \mathbb{R}^{n}$, and $\boldsymbol{g}_3 = \boldsymbol{g}_p \in \mathbb{R}^{n}$ generated as Perlin noise on the mesh \cite{perlin1985image, vigier_perlin_numpy}.
Note that the frequency of the functions $\boldsymbol{g}_i$ increases with $i$.
We define a patchwork function $\boldsymbol{q} \in \mathbb{R}^{n}$ on the mesh such that $\boldsymbol{q}[\text{group}_i] = \boldsymbol{g}_i[\text{group}_i]$. $\boldsymbol{y}_{rgb}$ is then derived by mapping $\boldsymbol{q}$, normalized to $[0,1]$, to a HSV colormap by defining the Hue parameter. See $\boldsymbol{y}_{rgb}$ in the GT figures in  Figure~\ref{fig:synthesis_comparison1}. 
%\vspace{-0.2cm}
\paragraph*{Results}
% \paragraph{\textbf{Results}}
Figure~\ref{fig:synthesis_comparison1} shows the error distributions for the fields learned by the three models, clipping errors above $5 \times 10^{-4}$ for clearer visualization.
Our N-Level model outperforms the others, exhibiting fewer artifacts. To provide quantitative results as well, Figure~\ref{fig:synthesis_comparison2} presents the Cumulative Distribution Functions (CDFs) of vertex errors across four groups: all vertices, $\text{group}_1$ (Red), $\text{group}_2$ (Yellowish), and $\text{group}_3$ (Blue), quantifying the percentage of vertices at each error level.
For each vertex $v$, error is measured as MSE: $\frac{1}{3} \lVert \boldsymbol{y}(v) - \boldsymbol{y}_{rgb}(v)\rVert_2^2$. The $x$-axis represents relative error, calculated by dividing vertex errors by the maximal error across all models for each subset. We note that the N-Level model shows superior performance for all groups except for the Red group, which corresponds to a constant function and is considered a trivial region.%\vspace{-0.2cm}
\subsection{Discontinuities and Exponential Scale Variations} \label{subsec:exper_2}
\begin{figure*}
  \centering  \includegraphics[width=0.8\textwidth]{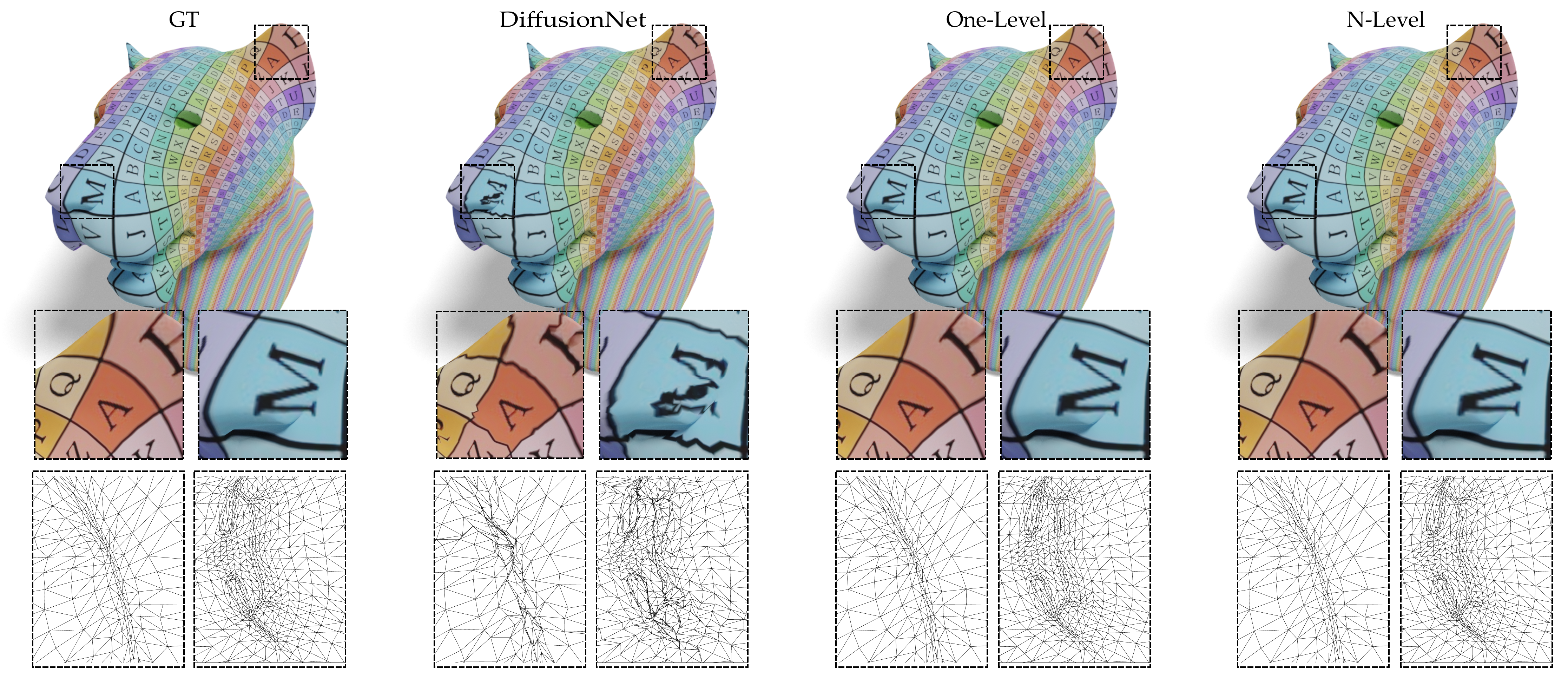}
  \caption{\textbf{Exponential scale variations.} From left to right: (1) GT texture, (2) Texture learned by the DiffusionNet model, (3) Texture learned by our One-Level model, (4) Texture learned by our N-Level model. The middle row zooms in on the texture in the nose and ear areas. The bottom row zooms in on the UV coordinates in the nose and ear areas. We observe that the DiffusionNet model exhibits significant texture distortions, while the textures learned by our One-Level and N-Level models closely resemble the GT texture.}
  \label{fig:uv_lion_texture}
\end{figure*}
\begin{figure*}
\centering
\includegraphics[width=0.85\textwidth]{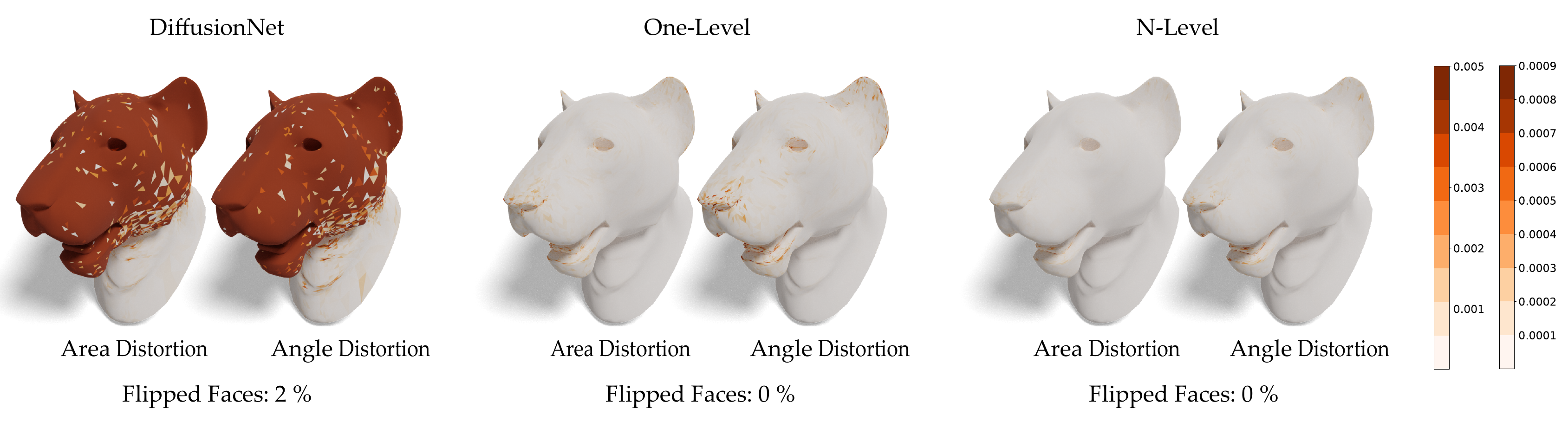}
\caption{\textbf{Exponential scale variations.} Three UV distortion metrics are presented for each model: area distortion, angle distortion, and the percentage of flipped faces. The area and angle distortion are visualized as mesh functions for each model, while the percentages of flipped faces are noted in text in the bottom row. Note that the DiffusionNet exhibits poor performance at the displayed error level, and that our N-level model attains the best results in terms of area and angle distortions.}
\label{fig:uv_distortions_lion}
\end{figure*}
We further evaluate our method on neural fields representing UV coordinates of textured meshes, which are typically non-continuous and exhibit exponential scale variations when generated by conformal parametrization.
% \change
% {
% Figure~\ref{fig:kangaroo-nffb} shows comparison between the UV learned by our N-Level model and the NFFB, and Figure~\ref{fig:uv_comparison_kangaroo} demonstrates our method's robustness by showing analysis of the learned UV coordinates by the different baselines.

% Figure~\ref{fig:uv_lion_texture} presents UV coordinates from a conformal map, highlighting our method's ability to handle exponential scale variations.
% We note that the while both our method and NFFB perform well on this smoother mapping, Figure~\ref{fig:kangaroo-nffb} demonstrates our method's handling of discontinuities.
% }

As in Section~\ref{subsec:exper_1}, we train the model by minimizing the mean square error (MSE), hence our loss function for this task is
\begin{equation}
\mathcal{L}_{\text{uv}}(\boldsymbol{y}) := \frac{1}{n} {\lVert \boldsymbol{y} - \boldsymbol{y}_{uv} \rVert }_{2}^{2}
\end{equation}
where $\boldsymbol{y}_{uv} \in \mathbb{R}^{n\times 2}$ defines the UV texture coordinates of vertices.%\vspace{-0.1cm}

\subsubsection{Discontinuity of Mesh and UV Coordinates}
\paragraph*{Data} In this example, we use a Kangaroo mesh with texture, with a total number of 10K vertices. The geometry of this mesh is composed of multiple connected components.
\paragraph*{Results}
Figure~\ref{fig:uv_comparison_kangaroo} displays the ground truth texture mesh (b) and error distributions for the three models (c, d, e), with errors above $1 \times 10^{-5}$ clipped for visualization.
The bottom row shows the 2D UV coordinates for each model. On the left (a), the CDFs of vertex errors are shown. Notably, DiffusionNet exhibits high errors at distinct features like eyes, nails, and tail tips. In its UV coordinates (g), DiffusionNet tends to squash and distort smaller regions.
The One-Level model, while showing significant errors at the head area, has less squashing than DiffusionNet but has distortions that cause overlaps with other texture components in UV (h).
Conversely, the N-Level model outperforms the others both qualitatively and quantitatively, with minimal distortions in its 2D UV coordinates, closely resembling the ground truth and with the best CDF results.

Figure~\ref{fig:uv_distortions_kangaroo} compares three UV distortion metrics for each model: area distortion, angle distortion, and the percentage of flipped faces.
Area distortion is measured by the absolute difference from 1 of the ratio between ground truth and predicted triangle areas, with values over 0.2 clipped. Angle distortion involves the absolute difference from 1 of the mean ratio between ground truth and predicted triangle angles, clipping values exceeding 0.01.
The bottom row notes in text the flipped faces percentages: 12\% for DiffusionNet, 16\% for One-Level, and 1\% for the N-Level model.
Overall, the N-Level model outperforms the others across all metrics.%\vspace{-0.1cm}

Figure~\ref{fig:kangaroo-nffb} shows a comparison between the UV learned by our N-Level model and NFFB. We observe that NFFB struggles particularly in regions where the texture mapping exhibits discontinuities, producing visible artifacts in these challenging areas. In contrast, our N-Level model generates a clean UV map without distortions.
Figure 2 in the supplemental material demonstrates this on an additional example, where in addition to difficulties with discontinuities the NFFB UV map exhibits noisy behavior in some regions.
% , and Figure~\ref{fig:uv_comparison_kangaroo} demonstrates our method's robustness by showing analysis of the learned UV coordinates by the different baselines.
% To further demonstrate the advantage of our architecture over state-of-the-art Euclidean methods, we compare our N-Level model with the NFFB architecture that inspired our work. 
% More details appear in the Supplemental Material.

\subsubsection{Exponential Scale Variations}
\paragraph*{Data}
In this example, we utilize a truncated lion head mesh with 8K vertices and UV coordinates computed using conformal mapping \cite{ben2008conformal}, which leads to exponential scale variations notably between the head and neck areas, see Figure~\ref{fig:uv_lion_texture}.%\vspace{-0.2cm}
\paragraph*{Results}
Figure~\ref{fig:uv_lion_texture} displays the GT texture alongside the results of the three models.
The top row features the textured mesh, while the middle and bottom rows show zoomed-in regions of the texture and UV coordinates, respectively.
The DiffusionNet texture reveals high distortion areas, but both the One-Level and N-Level models accurately capture the UV field.

Figure~\ref{fig:uv_distortions_lion} evaluates three UV distortion metrics for each model. Area distortion values above $5\times10^{-3}$ and angle distortion values above $9\times10^{-4}$ are clipped. The bottom row notes flipped face percentages: 2\% for DiffusionNet and 0\% for both the One-Level and N-Level models.
Although the One-Level and N-Level models show similar performance, the N-Level model still demonstrates the highest accuracy in area and angle distortions.%\vspace{-0.2cm}
We note that NFFB achieves results comparable to ours on this smoother mapping.
% We note that NFFB performs well on this smoother mapping.
% We note that the while that both our method and NFFB perform well on this smoother mapping, Figure~\ref{fig:kangaroo-nffb} demonstrates our method's handling of discontinuities.

\subsection{Mesh Generalization}
\label{subsec:exper_3}
\begin{figure}[h]
  \centering  \includegraphics[width=0.5\textwidth]{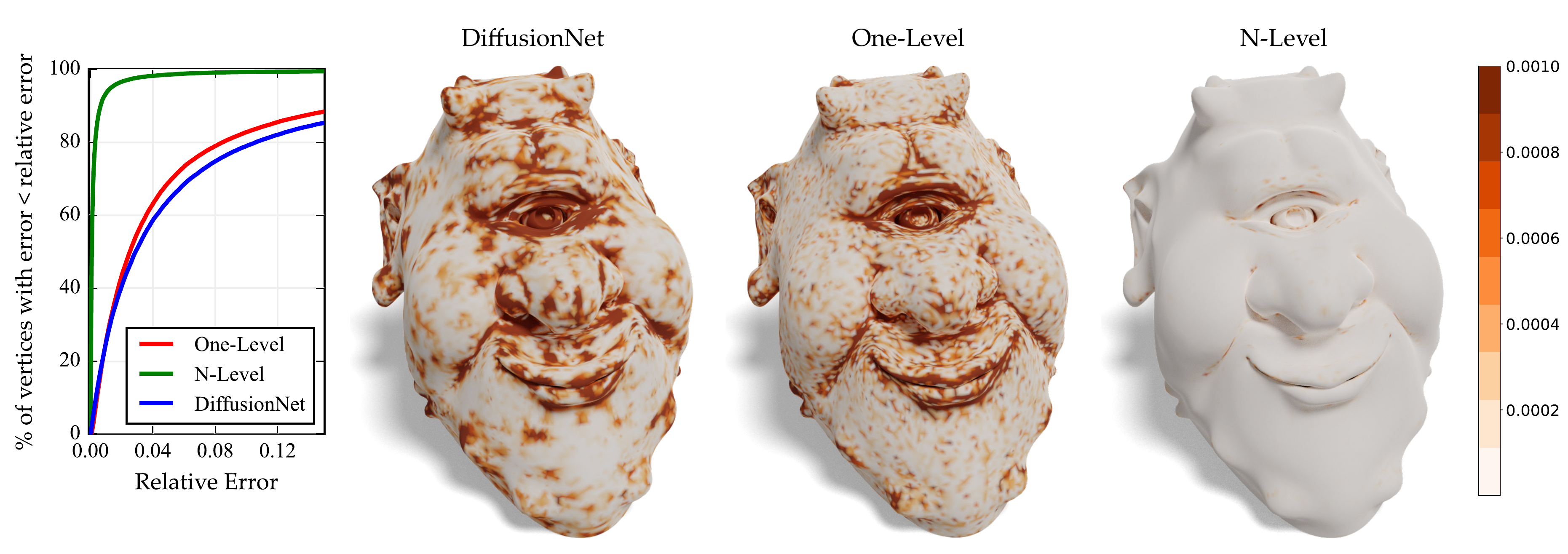}
  \caption{\textbf{Mesh Generalization.} From left to right: (1) CDFs of vertex errors for each of the three models. (2) Error distribution of the field learned by DiffusionNet. (3) Error distribution of the field learned by  our One-Level model. (4) Error distribution of the field learned by our N-Level model. Note that from both a quantitative and qualitative perspective, our N-Level model markedly surpasses the performance of the other two models.}
  \label{fig:comparison_ogre}
\end{figure}
\begin{figure}[h]
  \centering  \includegraphics[width=0.3\textwidth]{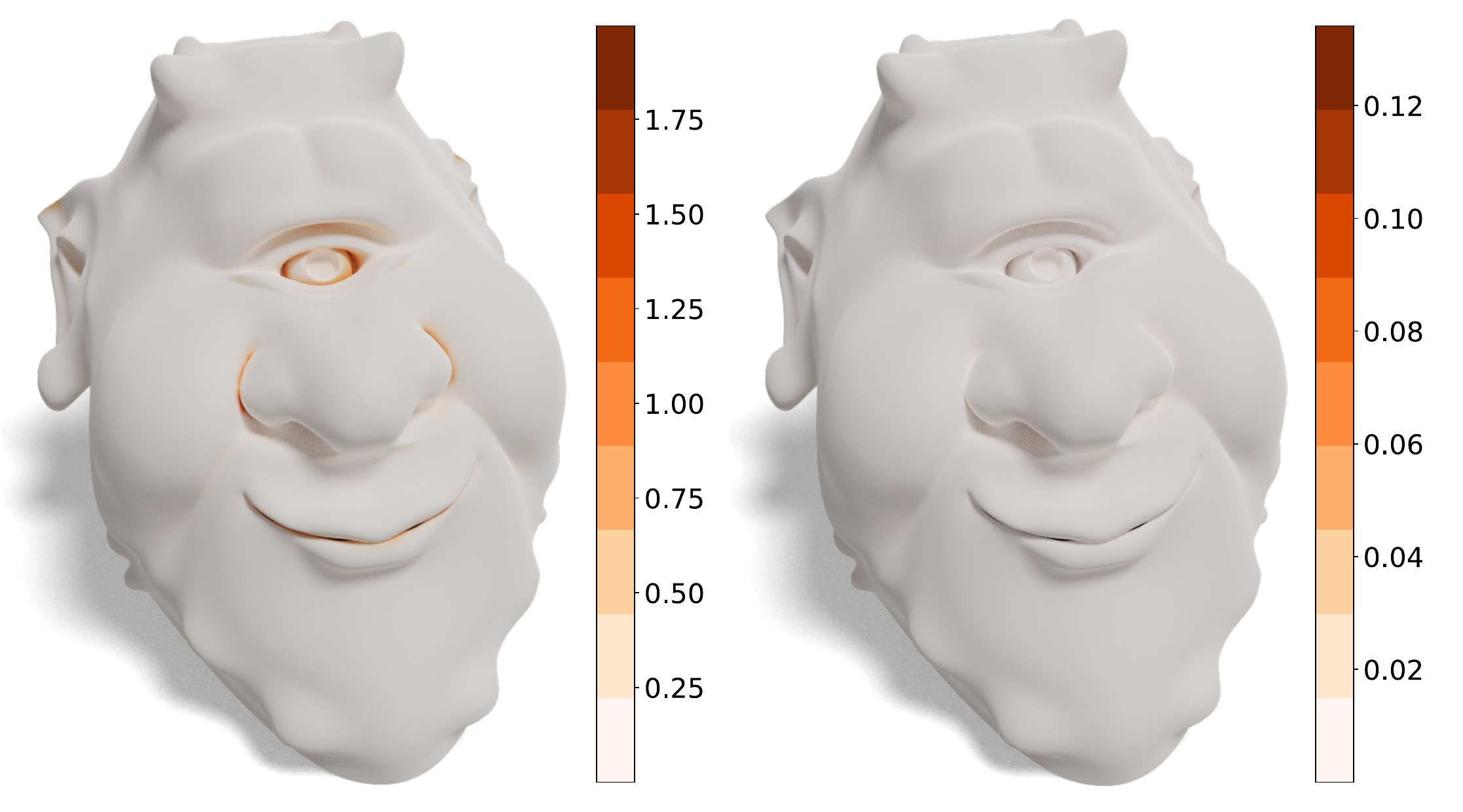}
  \caption{\textbf{Mesh Generalization.} Error comparison between our N-Level model and NFFB on test mesh. The NFFB was trained on the mesh that is one resolution level lower than the test. We observe that our model outperforms the NFFB. Note the different colorbars.}
  \label{fig:ogre_nffb}
\end{figure}
In this experiment, we demonstrate our architecture's ability to generalize across multiple versions of a single mesh. Starting with a base mesh, we generate several subdivided versions via a variant of Loop subdivision, as implemented by MeshLab \cite{MeshLab}, which avoids adding new vertices if triangle edges are below a specified threshold. We apply subdivision iterations until additional triangles are negligible. Since our base mesh is not overly coarse, not all triangles are subdivided in each iteration. However, before being fed into the network, each mesh is centered and normalized, making triangle additions significantly change the mesh embedding.

We aim to learn the neural field defined by mesh vertex normals, $\boldsymbol{y}_n \in \mathbb{R}^{n\times 3}$. Focusing on the direction of these normals, we train the model by minimizing the mean cosine distance error \cite{salton1975vector, han2022data}.
Thus, our loss function is given by:
\begin{equation}
\mathcal{L}_{\text{n}}(\boldsymbol{y}) := \frac{1}{n} \sum_{v}
\Bigg( {1 -  \frac{\langle \boldsymbol{y}(v), \boldsymbol{y}_n(v)  \rangle}{{\lVert \boldsymbol{y}(v) \rVert}_{2}^{2} \cdot {\lVert \boldsymbol{y}_n(v) \rVert}_{2}^{2}}
} \Bigg)
\end{equation}
where $\boldsymbol{y}_n(v) \in \mathbb{R}^3$ is the normal vector at vertex $v$.
\paragraph*{Data}
The base mesh used for the subdivision iterations is the smiling ogre mesh, which comprises of 20K vertices. We then generated five additional subdivided versions, with the largest containing 33K triangles and 65K faces. The dataset was split into training and testing sets; the training set comprises meshes subdivided through $[0,1,2,3,4]$ iterations, and the test set contains only the mesh generated by the 5-th subdivision iteration.%\vspace{-0.2cm}
% \paragraph{\textbf{Baselines}}
% We compare against the same baselines as in previous experiments: the DiffusionNet model and One-Level model.
\paragraph*{Results}
Figure~\ref{fig:comparison_ogre} displays the CDFs of vertex errors on the left, and the error distributions of the three models on the right, where clipping values exceeding $0.001$ for visualization. Both quantitatively and qualitatively, our N-Level model significantly outperforms the other two models.

Figure~\ref{fig:ogre_nffb} compares the error values on the test mesh between NFFB and our N-Level model. The visualization is shown without error clipping; note the different colorbar scales for each method. It can be observed that our model performs better. The NFFB was trained on a mesh that is one resolution level lower than the test mesh (4  subdivision iterations).

\section{Illustrative Application}
\label{sec:application}
\begin{figure*}[t]
    \centering
    \begin{subfigure}[t]{\linewidth}
        \centering
        \includegraphics[width=\linewidth]{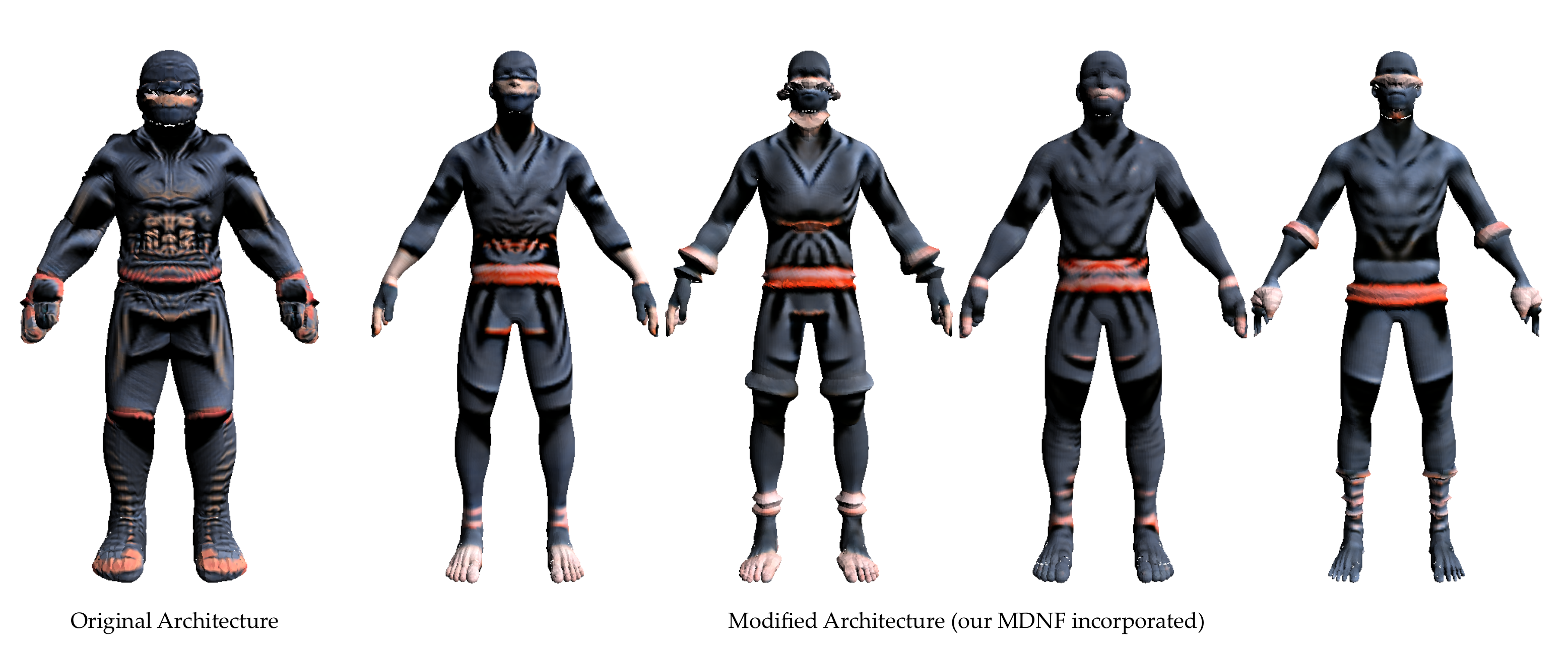}
        \caption{Prompt: \emph{"A 3D rendering of a ninja in unreal engine"}}
        \label{fig:text2mesh-ninja}
    \end{subfigure}
    \hfill
    \begin{subfigure}[t]{\linewidth}
        \centering
        \includegraphics[width=\linewidth]{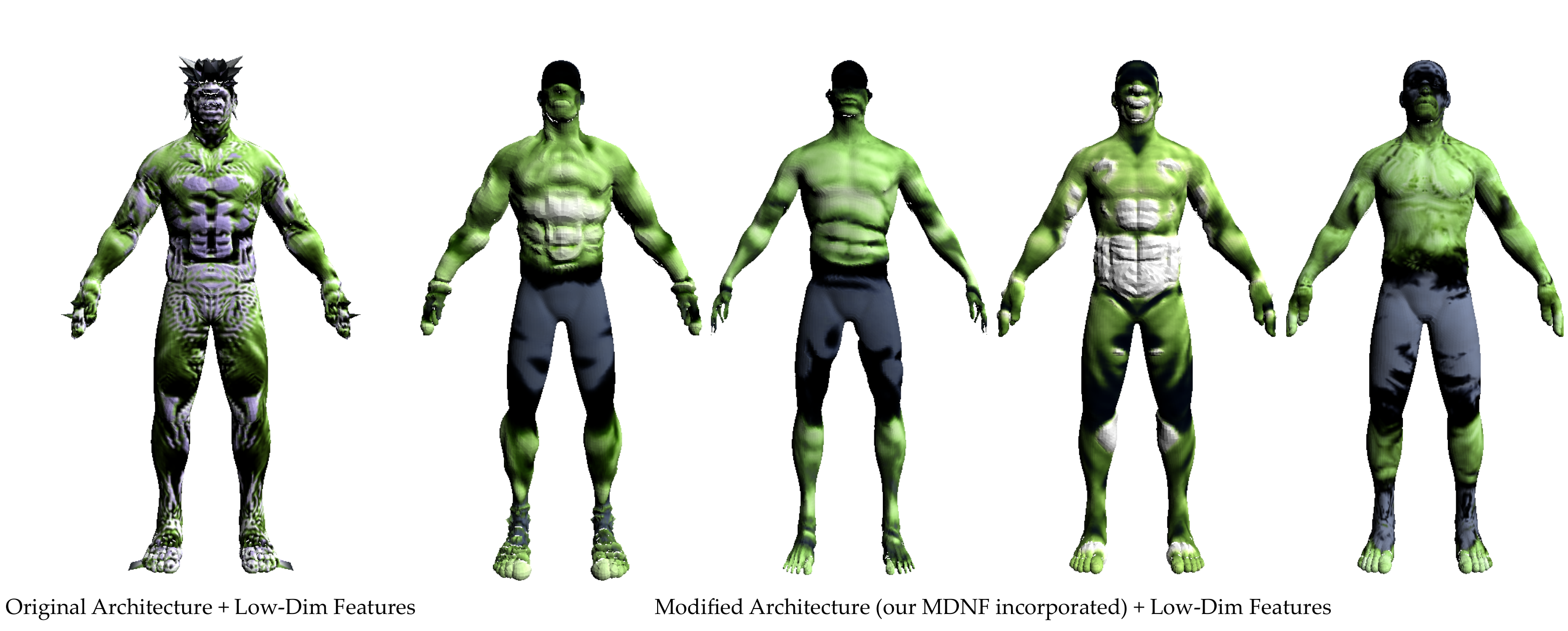}
        \caption{Prompt: "a 3D rendering of the Hulk in unreal engine"}
        \label{fig:text2mesh-hulk}
    \end{subfigure}
    \caption{\textbf{Illustrative Application.} Comparison between the text2mesh architecture without MDNF (left) and our modified version with MDNF (right) for the given prompts. See the experiments details in supplemental. While both models successfully generate meshes corresponding to the given prompts, our architecture achieves finer geometric resolution and enhanced detail preservation. Note the sharper features and more defined geometric details in our result, demonstrating the benefit of replacing the base MLP with our multi-resolution architecture.}
    \label{fig:text2mesh}
\end{figure*}
\begin{figure*}[t]
    \centering
    \includegraphics[width=\linewidth]{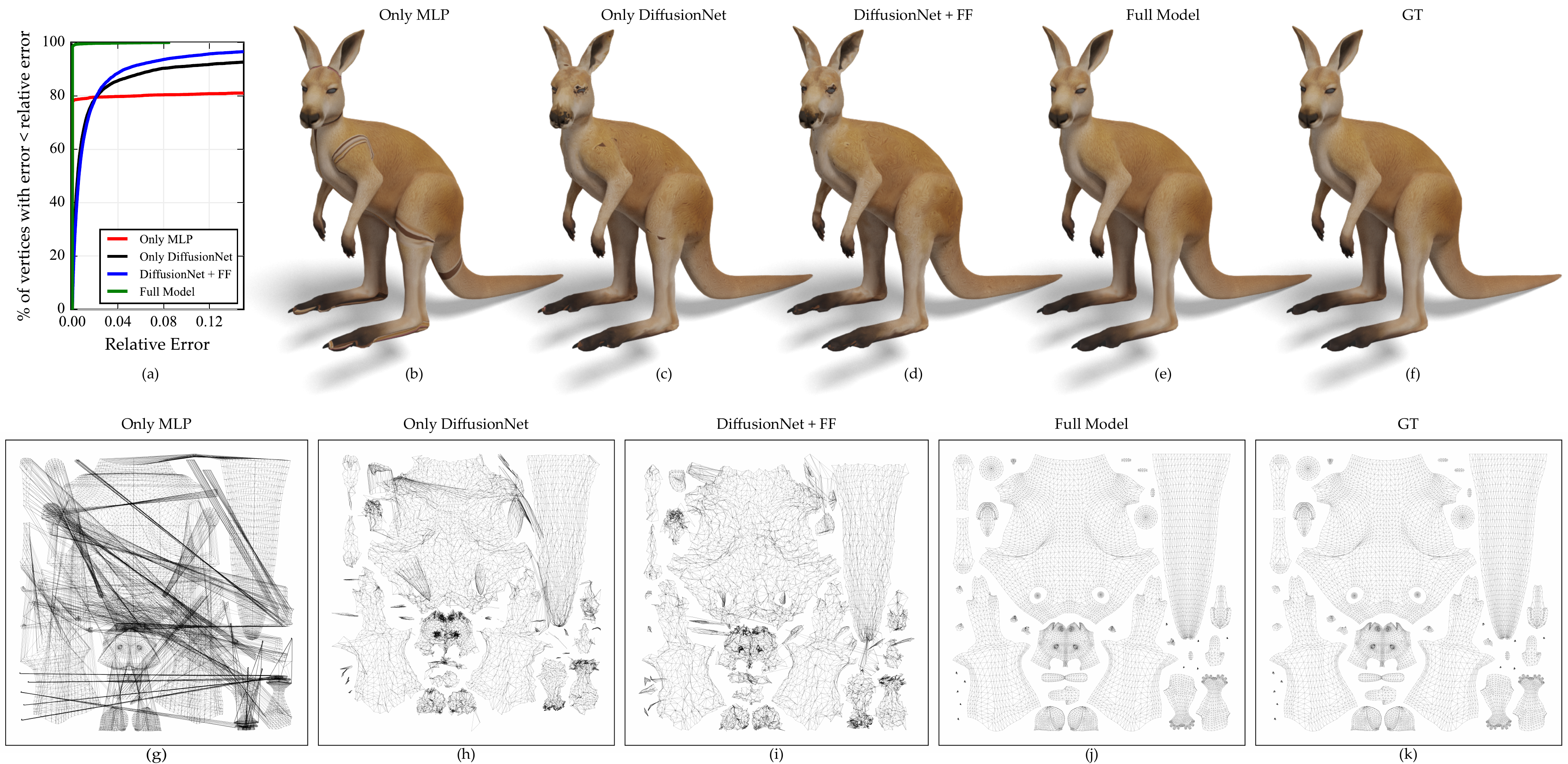}
    \caption{\textbf{Ablation Study.} Upper row: (a) Cumulative Distribution Functions (CDFs) of vertex errors for each of the three model variants and our full model. Texture learned by: (b) the "Only MLP" variant, (c) the "Only DiffusionNet" variant, (d) the "DiffusionNet + FF" variant, (e) our full model. (f) Ground truth textured mesh.
    Bottom row: Corresponding learned 2D UV coordinates for each texture. Note that our full model (e, j) is visually free from artifacts and achieves the best results both qualitatively and quantitatively, as shown in (a).
    }
    \label{fig:ablation_kangaroo}
\end{figure*}
% \begin{figure*}[t]
%     \centering
%     \includegraphics[width=0.9\linewidth]{figs/text2mesh_ninja_4ex.pdf}
%     \caption{\textbf{Illustrative Application.} Comparison between the original text2mesh architecture (left) and our modified version (right) for the prompt "a 3D rendering of a ninja in unreal engine". While both models successfully generate ninja-themed meshes, our architecture achieves finer geometric resolution and enhanced detail preservation. Note the sharper features and more defined geometric details in our result, demonstrating the benefit of replacing the base MLP with our multi-resolution architecture.}
%     \label{fig:text2mesh-ninja}
% \end{figure*}

To demonstrate the practical utility of our architecture, we integrate it into the text2mesh framework \cite{michel2022text2mesh}, a CLIP-based \cite{radford2021learning} model that generates textured meshes from text prompts. This architecture employs three MLPs, where a "base" MLP's output feeds into two subsequent networks that learn color and displacement values for each vertex of a base mesh.

In our experiment, we replace the base MLP with our architecture and compare the output of the original text2mesh implementation with our modified version output.
We reproduce two examples given in the paper, both using the same base human mesh.
The first example corresponds to the prompt \emph{"A 3D rendering of a ninja in unreal engine"} and the second example corresponds to the prompt \emph{"a 3D rendering of the Hulk in unreal engine"}.
We provide the details for both experiments in the supplemental material.
% In our experiment, we replace the base MLP with our architecture while maintaining all other components. Reproducing one of the examples given in the paper, we use the prompt \emph{"A 3D rendering of a ninja in unreal engine"} and the provided base human mesh, and compare the output of the original text2mesh implementation with our modified version output.
% We maintain the original parameters specified in \cite{michel2022text2mesh} while exploring various parameter configurations for our network module.

For consistency with the original implementation, we preserve text2mesh's zero initialization of weights and biases in the final layer of both color and displacement MLPs, which promotes neutral initial colors and displacements. We note that the final output is sensitive to this initialization.

Figure~\ref{fig:text2mesh-ninja} and Figure~\ref{fig:text2mesh-hulk} present results from both the original architecture and our modified version.
% Figure~\ref{fig:text2mesh-ninja} present results from both the original architecture and our modified version.
Across different parameter settings of our model, we observe a consistent trend: our architecture prioritizes producing meshes with finer geometric resolution, successfully capturing both high-frequency geometric details and visual features.
This enhanced detail preservation demonstrates our model's capacity to effectively represent multi-resolution signals.

\section{Ablation Study}
\label{sec:ablation}
% Beyond ablating the need of several components, the Illustrative example we provided in Figure~\ref{fig:method_dragon1} Section~\ref{sec:method} which shows the network output where different levels are disabled...?
To validate the design choices of our method, we conduct an ablation study.
Inspired by the original NFFB paper, we examine three variants: our method using only the DiffusionNet output, referred to as “Only DiffusionNet”; our method incorporating both DiffusionNet and Fourier feature encoding, denoted as “DiffusionNet + FF”; and a variant using only the sine-activated MLP, referred to as “Only MLP.”
For the “Only MLP” variant, we experimented with different network sizes to ensure a fair comparison in terms of model complexity.
Figure~\ref{fig:ablation_kangaroo} presents the results for the UV learning experiment with the kangaroo mesh model. The upper row illustrates the textured mesh learned by each variant (b, c, d) alongside our full model (e) and the ground truth (f). On the left (a), we show the cumulative distribution functions (CDFs) of vertex errors, and the bottom row shows the corresponding 2D UV coordinates.
We observe that our full model achieves superior results both qualitatively and quantitatively.

\section{Conclusions}
Our multi-resolution framework shows strong capability in representing neural fields on triangle meshes, achieving high precision across various domains and functions. Its detailed capture of fine features makes it ideal for high-precision tasks in computer graphics, such as UV learning, where a generally low error that suffices for applications like segmentation is not enough, and one needs to achieve close to machine precision. This framework can potentially be integrated into architectures addressing applications such as texture reconstruction from images, mesh stylization, and (as in the preliminary results that we have shown) texture and displacement generation.
%\change{Additionally, we provide an demonstration of its practical utility by integrating it into a texture generation model, suggesting its potential for preserving geometric fidelity.}
It also has the potential to serve as an effective feature extractor for other high-precision tasks in geometry processing.

%-------------------------------------------------------------------------
% bibtex
\bibliographystyle{eg-alpha-doi} 
\bibliography{egbibsample}       

% biblatex with biber
% \printbibliography                

%-------------------------------------------------------------------------
%Color tables are no longer required for purely electronic publications.
\newpage

\end{document}